\documentclass[a4paper]{cas-dc}

\usepackage[authoryear,longnamesfirst]{natbib}
\usepackage{adjustbox}
\def\tsc#1{\csdef{#1}{\textsc{\lowercase{#1}}\xspace}}
\tsc{WGM}
\tsc{QE}
\tsc{EP}
\tsc{PMS}
\tsc{BEC}
\tsc{DE}

\begin{document}
\let\WriteBookmarks\relax
\def\floatpagepagefraction{1}
\def\textpagefraction{0.001}
\shorttitle{}
\let\printorcid\relax

\title [mode = title]{GTA-Net: An IoT-Integrated 3D Human Pose Estimation System for Real-Time Adolescent Sports Posture Correction}

\author[1]{Shizhe Yuan}
\ead{ysz0920123@gmail.com}
\cormark[1]

\author[2]{Li Zhou}
\ead{li.zhou3@mail.mcgill.ca}

\address[1]{School of Physical Education, Xinyang Normal University, Xinyang , 464000, China
}

\address[2]{McGill University Montréal, 27708, Canada}

\begin{abstract}
With the advancement of artificial intelligence, 3D human pose estimation-based systems for sports training and posture correction have gained significant attention in adolescent sports. However, existing methods face challenges in handling complex movements, providing real-time feedback, and accommodating diverse postures, particularly with occlusions, rapid movements, and the resource constraints of Internet of Things (IoT) devices, making it difficult to balance accuracy and real-time performance. To address these issues, we propose GTA-Net, an intelligent system for posture correction and real-time feedback in adolescent sports, integrated within an IoT-enabled environment. This model enhances pose estimation in dynamic scenes by incorporating Graph Convolutional Networks (GCN), Temporal Convolutional Networks (TCN), and Hierarchical Attention mechanisms, achieving real-time correction through IoT devices. Experimental results show GTA-Net’s superior performance on Human3.6M, HumanEva-I, and MPI-INF-3DHP datasets, with Mean Per Joint Position Error (MPJPE) values of 32.2mm, 15.0mm, and 48.0mm, respectively, significantly outperforming existing methods. The model also demonstrates strong robustness in complex scenarios, maintaining high accuracy even with occlusions and rapid movements. This system enhances real-time posture correction and offers broad applications in intelligent sports and health management.
\end{abstract}


\maketitle

\section{Introduction}

In modern society, with the spread of health awareness and the promotion of educational policies, the proportion of adolescents participating in sports activities has significantly increased. Whether in school physical education classes or extracurricular sports clubs and training programs, young people are engaging in various sports to varying degrees~~\cite{nekoui2020falcons}. These activities not only contribute to the physical development and healthy growth of adolescents but also help cultivate their teamwork and competitive spirit. However, as sports activities become more widespread, another issue is gradually coming to the forefront—many adolescents, due to a lack of professional guidance, are prone to developing incorrect postures during exercise~\cite{mei20233d,ran2024brain}. While these incorrect postures may not cause obvious harm in the short term, over time, they can accumulate and lead to a series of sports injuries, such as muscle strains, arthritis, and scoliosis. These injuries not only affect the athletic performance of young people but could also pose serious threats to their physical development and long-term health~\cite{badiola2021systematic,kulkarni2021table}. Adolescents are in a crucial stage of physical development, and correct posture during exercise is vital to their growth~\cite{afsar2023body}. Unlike adults, adolescents' bones and muscles have not yet fully matured, and if incorrect postures are developed during this period, they may adversely affect the normal development of their skeletal structure and muscle groups, potentially leading to lifelong health problems. Therefore, it is particularly important to correct adolescents' postures in a timely manner and help them maintain the correct posture during exercise. However, traditional posture correction methods often rely on the experience of coaches or teachers, who judge students' postures through visual observation or video analysis~\cite{kwon2021youth}. This approach is not only time-consuming and labor-intensive but also susceptible to human error, leading to inconsistent correction results. Especially in large-scale physical education classes or training programs, it is challenging for coaches to monitor and correct each student's posture in real-time and with accuracy, resulting in many adolescents' incorrect postures going uncorrected.

With the continuous advancement of technology, the application of the Internet of Things (IoT) and Artificial Intelligence (AI) is becoming increasingly widespread in various fields, providing new technological means to address the issue of posture correction in adolescent sports~\cite{zhang2020empowering}. By utilizing advanced sensor technology and 3D human pose estimation algorithms, athletes' postures can be accurately captured and analyzed. Although these technologies have already been preliminarily applied in the training of some professional athletes, their application in the field of adolescent physical education remains relatively limited~\cite{xu2022vitpose}. Traditional correction methods are not only time-consuming and labor-intensive but also struggle to provide real-time feedback, failing to meet the needs of large groups of adolescents. Therefore, intelligent posture correction systems that integrate IoT technology have broad application prospects in adolescent physical education. These systems can use smart devices and sensor networks to monitor adolescents' postures in real-time and automatically generate correction suggestions, helping them adjust their posture during exercise promptly to avoid potential sports injuries~\cite{groos2021efficientpose}. The introduction of intelligent technology will make physical education more personalized and precise, effectively enhancing the scientific and safety aspects of adolescent sports training.

To address the challenges in adolescent posture correction during sports activities, early research predominantly utilized 2D image analysis for posture estimation. For instance, one study proposed a 2D posture estimation approach using deep neural networks (DNNs), which predicted posture by analyzing human key points frame by frame~\cite{dang2019deep}. This method processed images through multiple neural network layers to detect and locate body joints, performing well in static images. However, when dealing with dynamic movements, it struggled to track posture changes effectively due to a lack of temporal information, resulting in lower accuracy in complex scenarios, especially during fast, intricate motions.
To enhance posture estimation, other research employed convolutional neural networks (CNNs) to extract multi-scale features for key point detection~\cite{cao2023human}. While CNNs improved feature extraction and achieved high accuracy, their reliance on 2D information limited the ability to process 3D posture changes, making it challenging to handle continuous movements and occlusions.
In the realm of 3D posture estimation, researchers proposed a model based on fully connected networks (FCNs) that directly predicted 3D poses from 2D key points~\cite{ma2021context}. Despite its computational efficiency, the method’s reliance on high-quality input data reduced accuracy and stability when noise, occlusions, or changes in viewing angles were present, limiting its use in complex environments.
Another study explored 3D posture estimation using a single-view convolutional neural network~\cite{sun2020multi}. This approach simplified operations and reduced the need for multi-camera setups, showing good performance in simple movements. However, its accuracy was significantly constrained in complex scenarios, particularly with large rotations or occlusions, and the single viewpoint limited the ability to fully capture 3D motion complexity.


These studies highlight the fundamental limitations of previous approaches, such as the lack of temporal information, reliance on 2D data, and the inability to handle complex motion. Specifically, 2D posture estimation methods fall short in capturing dynamic changes and complex postures, while simple 3D estimation models are highly dependent on the quality of input data, making them susceptible to noise and occlusion. Additionally, in large-scale adolescent physical education settings, these methods struggle to provide efficient, low-latency real-time feedback, limiting their practical application. Therefore, there is an urgent need for a solution that combines 3D pose estimation, real-time feedback, and IoT integration to address these shortcomings and meet the demands of adolescent sports posture correction.

\begin{figure}
    \centering
	\includegraphics[width=0.5\textwidth]{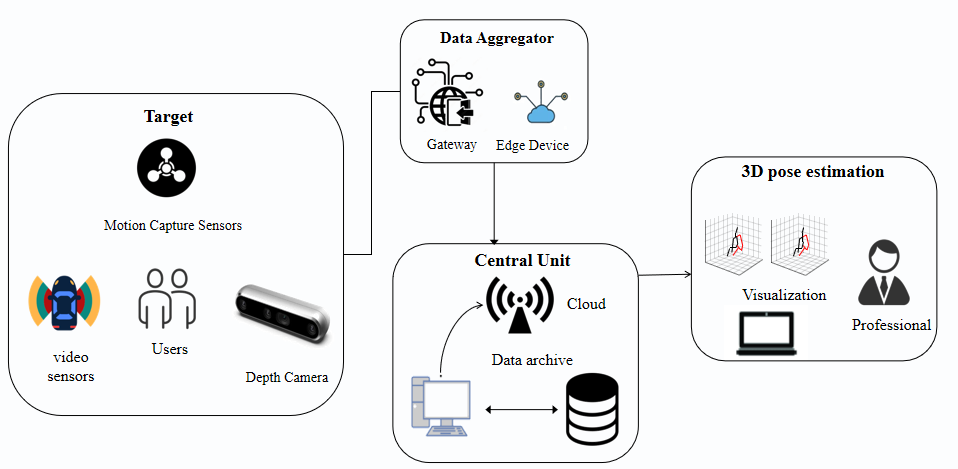}
    \caption{The IoT-based framework for 3D pose estimation and real-time feedback. The system collects posture data using motion capture sensors, video sensors, and depth cameras. This data is processed by a central unit via IoT technology, enabling real-time 3D pose estimation and feedback for posture correction.}
    \label{IOT}
\end{figure}

To address these challenges, we introduce GTA-Net, an innovative system designed for 3D human pose estimation, which not only accurately captures and analyzes adolescents' sports postures but also integrates Internet of Things (IoT) technology to enable real-time data transmission and feedback. Supported by IoT devices, our system can continuously monitor and analyze adolescents' postures in real-time without interfering with their normal physical activities. The data processing framework enabled by IoT technology is depicted in Figure \ref{IOT}, providing a typical example of how sensors, edge devices, and cloud computing work together for efficient data aggregation and pose estimation. It is important to note that this figure serves as a general reference for IoT-based data processing and not the overall architecture of our proposed GTA-Net system. Through the integration of advanced spatio-temporal modeling techniques such as Joint-GCN and Bone-GCN, and the incorporation of Attention-Augmented Temporal Convolutional Networks (TCN), the system ensures high-precision posture estimation across various sports scenarios while significantly improving data processing efficiency and feedback speed. These innovations allow our system to offer more robust posture analysis, addressing issues such as occlusions and complex movements that traditional methods struggle to handle. This system not only offers scientific support for posture correction among adolescents but also lays a solid foundation for preventing potential sports injuries. By introducing a novel architecture that combines 3D pose estimation with IoT for real-time monitoring and correction, this research provides key technological advancements that support the future development of intelligent sports systems and opens new pathways for scientifically managing adolescent sports activities.

Our contributions are primarily reflected in the following three aspects:

\begin{itemize}
\item We proposed an intelligent posture correction system based on 3D human pose estimation, enabling high-precision monitoring and correction of adolescents' sports postures. This overcomes the limitations of traditional 2D methods, which struggle to accurately capture dynamic and complex movements, ensuring more reliable and precise posture correction.
\item The system integrates IoT technology, achieving real-time data transmission and feedback, which significantly improves the system's responsiveness and user experience. Unlike previous methods that often suffer from latency, this system ensures timely feedback, which is crucial for immediate posture correction in sports training.
\item We validated the effectiveness and practicality of the system through experiments, providing a scientific solution for posture correction and injury prevention in adolescent sports. This contrasts with prior approaches that often lacked real-world validation, ensuring that our solution is both applicable and effective in practical sports training scenarios.
\end{itemize}

This research not only provides robust technical support for the scientific management of adolescent sports activities but also lays an important foundation for the future development of the intelligent sports field.

\section{Related Work}
\subsection{The Evolution from 2D to 3D Pose Estimation}

With the development of computer vision technology, human pose estimation has gradually evolved from the initial 2D pose estimation to 3D pose estimation. This evolution has not only improved the accuracy and applicability of pose estimation but has also enabled the technology to better adapt to complex real-world scenarios. The core of 2D pose estimation lies in detecting and locating human key points from single-view images, with these key points typically located at joints, limbs, and other critical positions, forming a 2D skeletal structure of the human body. Early methods primarily relied on traditional computer vision algorithms, such as the Hough Transform and Histogram of Oriented Gradients (HOG)~\cite{patel2020histogram}. As deep learning technology~\cite{caoapplication,luo2023fleet,chen2024enhancing,weng2024fortifying,xu2022dpmpc,peng2024maxk,xi2024enhancing} became more widespread, Convolutional Neural Networks (CNNs)~\cite{peng2023autorep,Wang2024Theoretical,qiao2024robust,liu2025eitnet,wang2024recording,luo2023aq2pnn,wang2024cross,sui2024application} gradually emerged as the mainstream approach for 2D pose estimation. Researchers developed various CNN-based models that achieved higher accuracy by detecting key points on feature maps at different resolutions~\cite{4zheng2023deep}. However, the limitation of 2D pose estimation is its inability to capture the posture information of the human body in 3D space, which makes it less effective in applications involving complex movements and depth changes~\cite{3wang2021deep}.
To address this limitation, researchers began exploring the transition from 2D to 3D pose estimation. One of the earliest attempts was mapping 2D key points to 3D coordinates to reconstruct 3D poses. Martinez et al. proposed a simple baseline model based on fully connected networks (FCNs) that regressed directly from 2D key points to predict 3D poses~\cite{martinez2019efficient}. The advantages of this method include its simple structure, high computational efficiency, and decent performance on standard datasets. However, it also has notable drawbacks, such as a high dependency on input data quality and poor accuracy and stability in the presence of noise, occlusion, and changes in viewing angles~\cite{5gholami2022self}. 
To overcome these challenges, GCN were introduced into 3D pose estimation. GCN effectively leverage the topological information of the human skeleton structure to map 2D key points into 3D space, enhancing the robustness of 3D pose estimation~\cite{6zhang2022semi}. Furthermore, researchers proposed methods based on Conditional Generative Adversarial Networks (Conditional GANs), which correct the generated 3D poses to further improve the accuracy and stability of the estimation~\cite{7zhang20233d}.

Although these methods have improved 3D pose estimation to some extent, they still face challenges when dealing with complex scenes and large movements. To address this, researchers introduced the concept of temporal dimensions, utilizing time-series analysis techniques to handle pose estimation across consecutive frames. Temporal Convolutional Networks (TCN) and Long Short-Term Memory Networks (LSTMs) have been widely used for such tasks, capturing the temporal dynamics of human actions to effectively improve the accuracy and coherence of 3D pose estimation~\cite{8gholami2022self}.
Additionally, the application of multi-view fusion technology has significantly driven the evolution from 2D to 3D pose estimation. Multi-view methods combine data from different camera viewpoints to reconstruct 3D poses more accurately~\cite{9luvizon2022consensus}. Sun et al. proposed a multi-view fusion model that integrates image information from multiple viewpoints, significantly enhancing the accuracy of 3D pose estimation~\cite{10sun2020multi}. However, the computational complexity of multi-view fusion is high, and finding ways to reduce computational costs while maintaining accuracy remains an important research direction.

In conclusion, the evolution from 2D to 3D pose estimation marks a significant leap in the field of pose estimation. This evolution has not only expanded the application scenarios of pose estimation but has also driven the development of more complex and accurate algorithms. Although current 3D pose estimation methods have made remarkable progress, many challenges remain, particularly in terms of real-time performance and large-scale applications. These challenges point the way for future research.

\subsection{Applications and Technical Challenges of 3D Pose Estimation in Complex Scenarios}

3D pose estimation technology faces numerous technical challenges when applied to complex real-world scenarios. Complex scenarios refer to environments involving multiple targets, multi-view perspectives, rapid motion, large rotations, and occlusions. These challenges demand that pose estimation algorithms not only achieve high accuracy but also find a balance between computational efficiency and real-time performance. Despite significant progress in recent years, there are still many difficulties in applying these technologies effectively in such complex environments.

The complexity of 3D pose estimation increases significantly in multi-target detection and interaction scenarios. Traditional single-person pose estimation methods often perform poorly in multi-target environments, particularly when multiple objects occlude each other or are densely arranged~\cite{21fang2022alphapose,10603432}. To address this issue, researchers have developed multi-target pose estimation methods, which often combine multi-view data fusion with scene context information to distinguish between different targets~\cite{22chen2024multi}. For instance, Cheng et al. proposed a distance-aware multi-target 3D pose estimation method that maintains high detection accuracy in dense scenes~\cite{23cheng2021monocular}. However, real-time tracking and estimation of multiple objects' poses in dynamic scenarios remain a challenge that requires further research.
3D pose estimation in scenarios involving rapid motion and large rotations is equally challenging. In such scenarios, the human body's pose changes quickly and significantly, and traditional pose estimation algorithms may fail to capture these changes in time, leading to increased estimation errors~\cite{24zhou2020tracking}. To enhance robustness, researchers have explored techniques that combine optical flow methods with multi-frame information fusion, analyzing motion trajectories across several adjacent frames to improve stability and accuracy without significantly increasing computational overhead~\cite{25benzine2021single,26kadkhodamohammadi2021generalizable, gong2024graphicalstructurallearningrsfmri,dong2024design,zhang2024deep,wang2018performance,liu2025eitnet,chen2024mix,wan2024image,weng2024big}. However, these techniques still face challenges in practical applications, particularly in maintaining high precision while reducing computational complexity.

Occlusion is another critical technical challenge for 3D pose estimation in complex scenarios. In multi-person interaction or crowded scenes, occlusion becomes particularly pronounced, with some key points potentially undetectable due to occlusion, thereby affecting overall pose estimation performance~\cite{27xia20203d}. To address this problem, researchers have introduced context-aware and partial reconstruction techniques, which infer and complete missing key points based on surrounding context, significantly improving the robustness of pose estimation~\cite{28nogueira2024markerless, cao2018expected,wang2024intelligent,chen2024few,peng2024automatic,weng2024leveraging,zheng2024identification,Shen2024Harnessing,zhou2024adapi,jin2024learning}. Additionally, multi-view fusion technology has been widely applied to solve occlusion issues by integrating image data from different perspectives, allowing better reconstruction of occluded key points and improving overall estimation accuracy~\cite{29maaz2022edgenext}.
Balancing real-time performance with computational efficiency is also crucial in the practical application of 3D pose estimation. In scenarios requiring high dynamic response, achieving real-time processing while maintaining estimation accuracy is a significant challenge for researchers. Although many algorithm optimization strategies have emerged in recent years, such as model pruning, quantization, and knowledge distillation techniques to reduce computational load~\cite{30nowak2021weight}, these methods often struggle to significantly improve real-time performance without sacrificing accuracy. In this regard, future research may need to explore new balances between hardware acceleration, algorithm optimization, and efficient data processing techniques to achieve broader application~\cite{31nowak2021weight, zhou2024optimization,liu2024dsem,li2024deep,yan2024application,huang2024risk,li2024optimizing,wang2024deep,zhang2024cu}.
Moreover, the challenges of 3D pose estimation vary across different application domains. For example, in Virtual Reality (VR) and Augmented Reality (AR) applications, pose estimation requires not only high accuracy but also tight synchronization with virtual environments, imposing higher demands on real-time performance and computational efficiency~\cite{32postolache2020remote}. In the field of autonomous driving, 3D pose estimation technology must quickly recognize and process various pedestrian poses in complex road environments to ensure driving safety~\cite{33kim2020road}. These differing requirements across fields further complicate technological development, and future research should focus more on the unique challenges within these specific applications.

In summary, the application of 3D pose estimation in complex scenarios faces multiple challenges, including multi-target detection, rapid motion, occlusion, and the balance between real-time performance and computational efficiency. While significant technological advances have been made, further research and innovation are needed to solve these problems and ensure that 3D pose estimation technology can realize its full potential in more practical applications.

\section{Method}
\subsection{Overview of our network}

\begin{figure*}
    \centering
    \includegraphics[width=1\linewidth]{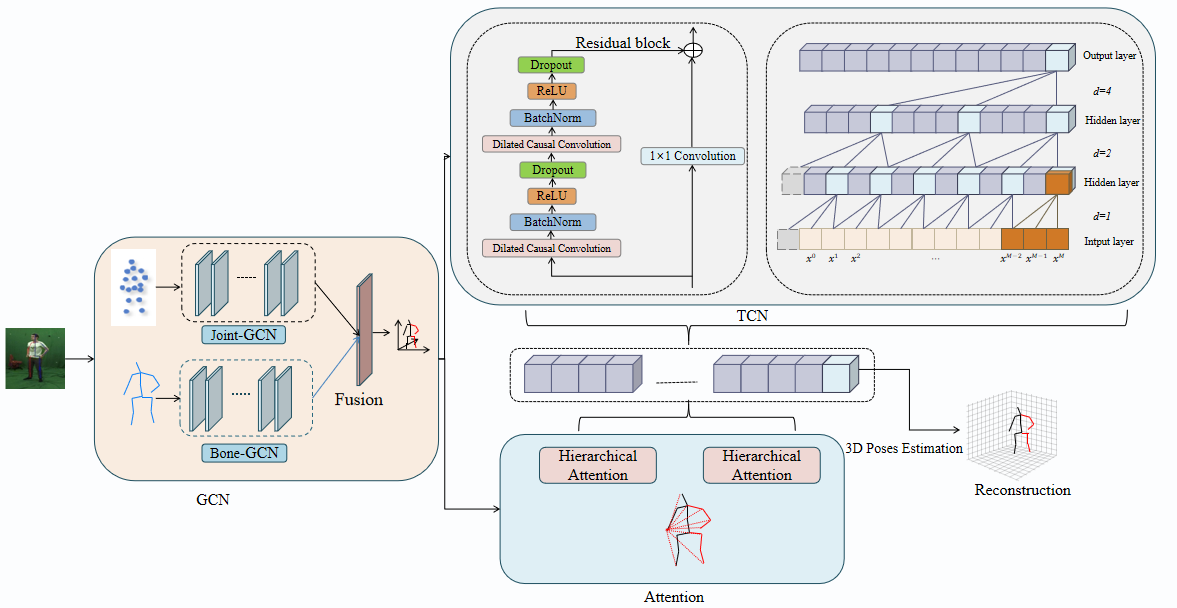}
    \caption{The overall architecture of the proposed system, combining GCN, TCN, and Hierarchical Attention mechanisms.The system utilizes Joint-GCN and Bone-GCN to capture local and global spatial relationships, while the Hierarchical Attention-augmented TCN refines temporal dynamics, enabling accurate 3D pose estimation in dynamic sports scenarios.}
    \label{overall}
\end{figure*}

Considering the limitations of previous studies in complex 3D posture estimation, we propose a model called GTA-Net (Graph-Temporal-Attention Network), specifically designed for posture correction and real-time feedback in youth sports. Our model leverages the strengths of GCN, TCN, and attention mechanisms to address the challenges inherent in dynamic and complex motion scenarios. The overall architecture of GTA-Net is depicted in Figure \ref{overall}, illustrating how these components are integrated within the model. 

GTA-Net consists of three primary components: Joint-GCN, Bone-GCN, and a Hierarchical Attention-augmented TCN. The process begins with Joint-GCN, which captures intricate relationships between individual joints by processing the 2D keypoints derived from video frames. This module focuses on extracting local spatial features related to each joint. On the other hand, following the Joint-GCN, the Bone-GCN focuses on the relationships between the bones that connect these joints, capturing the global structural information of the human body. By processing the skeletal connections, Bone-GCN ensures that the overall pose estimation remains consistent and stable, even in complex motion scenarios. The integration of these two components allows GTA-Net to effectively leverage both local and global context, leading to a more accurate and coherent 3D pose estimation.
Once the 3D keypoint features are extracted from the GCN modules, they are subsequently fed into the next stage: the Hierarchical Attention-augmented TCN. This TCN is designed to handle the temporal dynamics of human motion, capturing continuous changes in joint positions over time. To refine this temporal modeling, the Hierarchical Attention mechanism, which includes both Temporal Attention and Spatial Attention layers, is integrated into the TCN. The Temporal Attention layer dynamically assigns weights to different time frames, allowing the model to prioritize frames that have a greater impact on overall pose estimation. Meanwhile, the Spatial Attention layer enhances the model's ability to focus on key spatial features, ensuring that the most relevant joints and bones are given higher priority in the estimation process. This hierarchical attention approach not only improves the model's accuracy but also enhances its robustness and generalization across different motion patterns.

The network is implemented within a framework that supports real-time feedback by integrating IoT technologies for seamless data transmission and processing. The IoT infrastructure facilitates the collection and transfer of data from multiple devices, such as cameras and sensors, to the processing units where the GTA-Net model operates.  This distributed approach ensures that the system can provide immediate corrective feedback to users, which is crucial for applications like sports training and posture correction, where real-time guidance is essential.

GTA-Net overcomes the limitations of 3D pose estimation in dynamic environments by integrating joint and bone information with a temporal attention mechanism, ensuring precise and consistent results even with rapid movements. Its IoT-enabled framework allows for real-time feedback, making it particularly effective for posture correction in adolescent sports.
\subsection{Joint-GCN and Bone-GCN}

Joint-GCN and Bone-GCN are two variants of the Graph Convolutional Network (GCN), specifically designed to capture the complex spatial relationships in human pose estimation. Joint-GCN primarily handles the local relationships between joints, modeling and extracting features of each joint through graph convolution operations. In the task of human pose estimation, each joint (such as shoulders, elbows, knees, etc.) has complex dependencies with its adjacent joints. Joint-GCN builds a graph structure between joints and utilizes graph convolutional layers to propagate and update features on this structure, effectively capturing these local spatial relationships. This allows Joint-GCN to precisely model details of posture changes, such as minor movements of joints, which is crucial for identifying and correcting subtle errors in posture. On the other hand, Bone-GCN focuses on the skeletal connections between joints. The skeleton, composed of multiple joints, represents the overall structure of the human body. Bone-GCN captures the global posture information by modeling the features of these skeletal connections. Unlike Joint-GCN, Bone-GCN emphasizes the stability and consistency of the overall structure, ensuring that the model maintains an accurate understanding of the posture even during rapid or extensive movements. Bone-GCN introduces a graph structure of skeletal connections between joints, propagating and integrating features to capture global posture information, thereby enhancing the model's ability to handle complex movements.

The overall architecture is shown in Figure \ref{GCN}, which displays the integration and roles of the Joint-GCN and Bone-GCN modules within the GTA-Net. The working principles of Joint-GCN and Bone-GCN in the graph convolutional network can be expressed through the following mathematical formulas:

\begin{figure*}[h]
	\centering
	\includegraphics[width=0.8\textwidth, height=0.5\textheight]{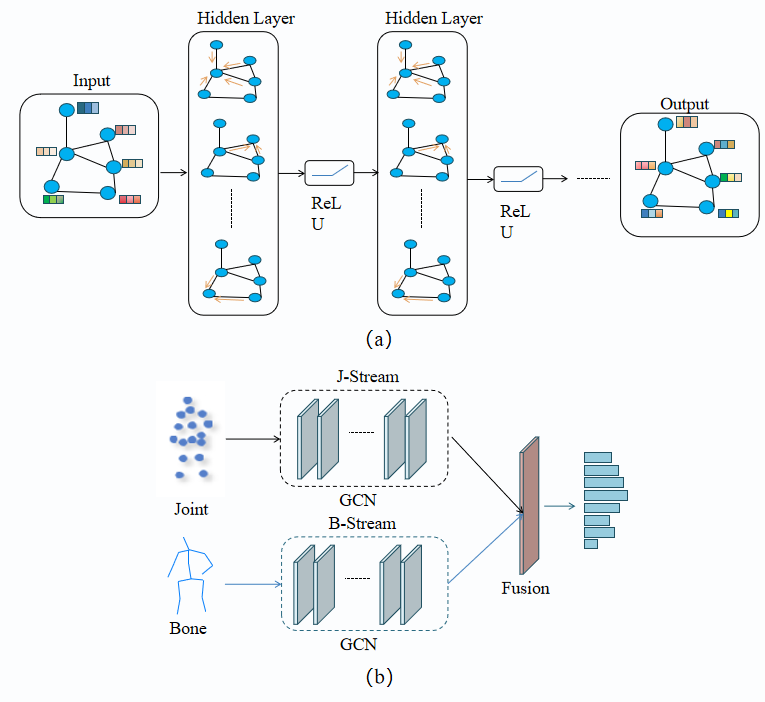}
	\caption{The structure of the proposed GCN model: (a) illustrates the basic Graph Convolutional Network operation, while (b) presents the dual-stream architecture combining joint (J-Stream) and bone (B-Stream) information for enhanced 3D pose estimation. The integration of Joint-GCN and Bone-GCN modules improves the system's ability to capture both local and global spatial relationships, essential for accurate pose estimation.}
	\label{GCN}
\end{figure*}

GCN Layer Operation: The fundamental operation in a GCN involves updating the node features by aggregating information from neighboring nodes. This is mathematically represented as:

\begin{equation}
H^{(l+1)} = \sigma\left( D^{-\frac{1}{2}} A D^{-\frac{1}{2}} H^{(l)} W^{(l)} \right)
\end{equation}
where $H^{(l)}$ is the feature matrix at layer $l$, $A$ is the adjacency matrix, $D$ is the degree matrix, $W^{(l)}$ is the trainable weight matrix at layer $l$, and $\sigma$ is the activation function.

Adjacency Matrix with Self-Loops: To ensure that a node can retain its own features during the aggregation process, self-loops are added to the adjacency matrix:

\begin{equation}
A' = A + I
\end{equation}
where $A'$ is the adjacency matrix with added self-loops, $A$ is the original adjacency matrix, and $I$ is the identity matrix.

Normalization of Adjacency Matrix: The adjacency matrix with self-loops is then normalized to stabilize the learning process and maintain consistent node importance:

\begin{equation}
\hat{A} = D^{-\frac{1}{2}} A' D^{-\frac{1}{2}}
\end{equation}
where $\hat{A}$ is the normalized adjacency matrix, $D$ is the degree matrix derived from $A'$, and $A'$ is the adjacency matrix with self-loops.

Feature Propagation: The node features are propagated through the network using the normalized adjacency matrix, resulting in the output feature matrix after the first GCN layer:

\begin{equation}
Z = \hat{A} H^{(0)} W^{(0)}
\end{equation}
where $Z$ is the output feature matrix after the first GCN layer, $\hat{A}$ is the normalized adjacency matrix, $H^{(0)}$ is the input feature matrix, and $W^{(0)}$ is the trainable weight matrix for the first layer.

Loss Function: The cross-entropy loss function is used to optimize the model during training, comparing the predicted labels with the ground truth:

\begin{equation}
\mathcal{L} = - \sum_{i \in \mathcal{Y}_L} \sum_{c=1}^{C} Y_{ic} \log Z_{ic}
\end{equation}
where $\mathcal{L}$ is the cross-entropy loss, $\mathcal{Y}_L$ is the set of labeled nodes, $C$ is the number of classes, $Y_{ic}$ is the label indicator for node $i$ and class $c$, and $Z_{ic}$ is the predicted probability for node $i$ and class $c$.

The integration of Joint-GCN and Bone-GCN provides complementary spatial feature extraction, enhancing the accuracy and stability of pose estimation. Joint-GCN processes the detailed relationships between joints, ensuring high accuracy even for subtle and complex motion changes. This is particularly important in adolescent sports training, where minor posture deviations can lead to injuries. Bone-GCN’s modeling of the global skeletal structure enhances stability in posture estimation, preventing overall posture distortion. The collaborative operation of these two modules enables GTA-Net to capture fine-grained details while maintaining coherence during dynamic movements.

\subsection{Temporal Convolutional Network}

\begin{figure}[h]
	\centering
	\includegraphics[width=0.5\textwidth]{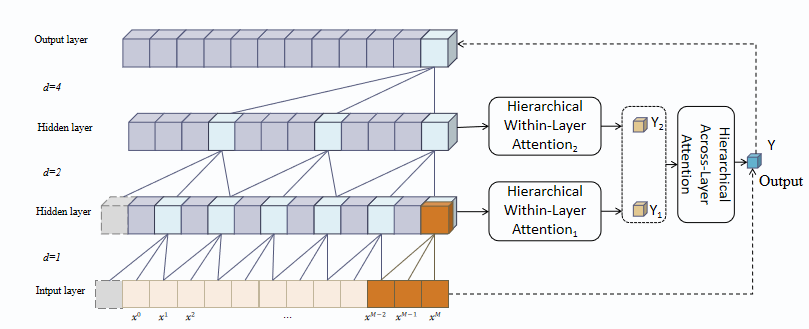}
	\caption{The structure of the Attention-Augmented Temporal Convolutional Network.This network is responsible for capturing the temporal dynamics of human motion, with attention mechanisms refining temporal features to improve pose estimation in dynamic movements. The TCN applies causal and dilated convolutions to efficiently process time-series data, ensuring real-time performance.}
	\label{TCN}
\end{figure}

The Attention-Augmented TCN is a neural network architecture specifically designed for processing time-series data, enhanced with attention mechanisms to improve the model's focus on important temporal features. Unlike traditional Recurrent Neural Networks (RNNs), the TCN captures temporal dependencies in sequential data through one-dimensional convolution operations, resulting in significant advantages in computational efficiency and model performance. A key feature of the TCN is its causal convolution, which ensures that the output at each time step depends only on the current and previous inputs, preserving the sequence order of the time-series data. This causal convolution design prevents future information from influencing the current state, ensuring temporal consistency in model inference. Additionally, TCN uses dilated convolutions to capture dependencies over longer time ranges without increasing network depth, enabling it to effectively model complex temporal dynamics in long time-series data. The integration of Hierarchical Within-Layer and Across-Layer Attention mechanisms further refines the temporal features, enhancing the model's ability to capture complex temporal patterns. The working principle of the Attention-Augmented TCN is illustrated in Figure \ref{TCN}, which shows its operation in processing time-series data.


In the GTA-Net model, TCN is used to capture and process the temporal dynamics of human motion. By applying layer-by-layer convolutions, TCN transforms the input time-series data into high-dimensional feature representations, allowing the model to accurately track the continuous changes in joint positions during motion. These temporal features are crucial for pose estimation, especially when dealing with complex motions in long time-series data. TCN identifies and maintains the temporal dependencies of the pose, providing critical support for the model's pose correction and real-time feedback. The causal convolution characteristic of TCN ensures consistency in pose estimation across the temporal dimension, effectively preventing future information from interfering with the current pose estimation. This ability to capture dynamic temporal information significantly enhances the accuracy and real-time performance of pose correction. To better understand TCN's specific operations within GTA-Net, the following equations describe the key computational steps involved in modeling temporal dynamics.

Causal Convolution: The core operation in a TCN involves applying causal convolutions to ensure that each output depends only on the current and past inputs, maintaining the temporal order.

\begin{equation}
y_t = \sum_{i=0}^{k-1} w_i \cdot x_{t-i}
\end{equation}
where $y_t$ is the output at time step $t$, $w_i$ are the convolution filter weights, $x_{t-i}$ represents the input at time step $t-i$, and $k$ is the filter size.

Dilated Convolution: To capture long-range dependencies without increasing the network depth, dilated convolutions are used, which introduce a dilation factor $d$.

\begin{equation}
y_t = \sum_{i=0}^{k-1} w_i \cdot x_{t-i \cdot d}
\end{equation}
where $d$ is the dilation factor, and other variables are defined as above.

Temporal Convolution Layer: The temporal convolution operation for the entire layer can be represented as a transformation of the input sequence $X$ into an output sequence $Y$.

\begin{equation}
Y = W \ast_d X
\end{equation}
where $Y$ is the output sequence, $W$ represents the convolutional filter weights, $X$ is the input sequence, and $\ast_d$ denotes the dilated convolution operation.

Residual Connection: TCN often include residual connections to improve gradient flow during training, where the output is combined with the input.

\begin{equation}
Z = Y + X
\end{equation}
where $Z$ is the final output after the residual connection, $Y$ is the output from the convolutional layer, and $X$ is the original input.

Loss Function: The model is trained by minimizing the loss function, typically the mean squared error (MSE) for regression tasks.

\begin{equation}
\mathcal{L} = \frac{1}{N} \sum_{t=1}^{N} (Z_t - \hat{Y}_t)^2
\end{equation}
where $\mathcal{L}$ is the loss, $N$ is the number of time steps, $Z_t$ is the predicted output, and $\hat{Y}_t$ is the ground truth at time step $t$.

These formulas sequentially describe the operations in a TCN, starting from the basic causal convolution, extending to dilated convolutions, and concluding with the inclusion of residual connections and the loss function used for training. Each formula builds upon the previous one to provide a comprehensive understanding of how TCN processes temporal data.

\subsection{Hierarchical Attention Mechanism}

\begin{figure}
	\centering
	\includegraphics[width=0.5\textwidth]{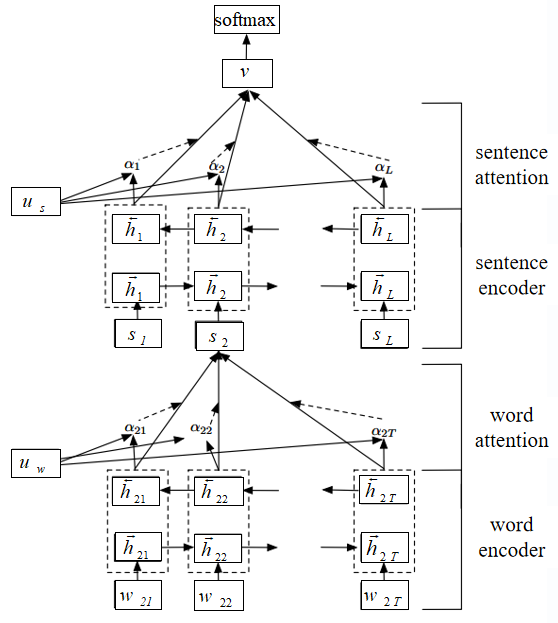}
	\caption{The hierarchical attention network architecture. \cite{yang2016hierarchical}. This architecture integrates Hierarchical Attention into the GCN-TCN framework, allowing the model to focus on critical features across different levels, from local joints to global skeletal structures. This multi-level attention ensures more precise and stable 3D pose estimation in complex motion scenarios.}
	\label{attention}
\end{figure}

Hierarchical Attention is a powerful technique in deep learning that dynamically adjusts focus across different levels of a hierarchical structure to capture and process the most important features in the data. Initially applied in natural language processing tasks, this mechanism allocates attention weights across different levels, such as words, sentences, and paragraphs, allowing the model to focus on the most critical information. The core idea is to enhance the model’s understanding of both contextual and global semantics through hierarchical attention, thereby improving the extraction of valuable feature information. This multi-level aggregation enables the model to better comprehend and leverage the semantic relationships within complex data, ultimately enhancing its performance. The structure and operation of this attention mechanism, as implemented in our TCN, are illustrated in Figure \ref{attention}, highlighting how Hierarchical Within-Layer and Across-Layer Attention are applied to refine temporal features for more accurate predictions.

The following key mathematical equations outline the Hierarchical Attention mechanism as applied in our model, with each equation explained in detail:

Hidden Representation:
The hidden representation $h_i$ is computed using a linear transformation followed by a ReLU activation function.
\begin{equation}
h_i = \text{ReLU}(W_h x_i + b_h)
\end{equation}
where $h_i$ represents the hidden representation of the input feature $x_i$, $W_h$ is the weight matrix, and $b_h$ is the bias term.

Attention Score Calculation:
The attention score $e_{ij}$ between nodes $i$ and $j$ is calculated using the dot product of their transformed representations, scaled by the square root of the dimension of the key vector.
\begin{equation}
e_{ij} = \text{attn}(h_i, h_j) = \frac{(h_i W_q)(h_j W_k)^T}{\sqrt{d_k}}
\end{equation}
where $e_{ij}$ is the attention score between nodes $i$ and $j$, $W_q$ and $W_k$ are the query and key weight matrices, respectively, and $d_k$ is the dimension of the key vector.

Attention Weight Normalization:
The attention weight $\alpha_{ij}$ is obtained by normalizing the attention score $e_{ij}$ using the softmax function over all neighbors of node $i$.
\begin{equation}
\alpha_{ij} = \frac{\exp(e_{ij})}{\sum_{k \in \mathcal{N}(i)} \exp(e_{ik})}
\end{equation}
where $\alpha_{ij}$ is the normalized attention weight for node $j$ relative to node $i$, and $\mathcal{N}(i)$ represents the neighbors of node $i$.

Feature Aggregation:
The updated node representation $h'_i$ is computed by aggregating the weighted features from its neighbors, using the attention weights $\alpha_{ij}$.
\begin{equation}
h'_i = \sum_{j \in \mathcal{N}(i)} \alpha_{ij} h_j
\end{equation}
where $h'_i$ is the updated representation of node $i$ after aggregating the weighted features from its neighbors $j \in \mathcal{N}(i)$, based on the attention weights $\alpha_{ij}$.

Final Aggregated Representation:
The final representation $z$ for the entire graph or sequence is obtained by summing the updated node representations and applying a linear transformation followed by a ReLU activation function.
\begin{equation}
z = \text{ReLU}(W_z \sum_{i=1}^{N} h'_i) + b_z
\end{equation}
where $z$ represents the final aggregated representation for the entire graph or sequence, $W_z$ is the weight matrix, $N$ is the number of nodes or time steps, and $b_z$ is the bias term.

In our GTA-Net model, we innovatively integrate Hierarchical Attention into the complex task of 3D posture estimation to address its diverse challenges. Hierarchical Attention first allows the model to flexibly adjust its focus at the joint and skeletal levels, enabling GTA-Net to more precisely model the spatial and temporal relationships within human posture. This approach not only helps the model identify and analyze subtle changes in individual joints but also integrates these spatial relationships for a more comprehensive understanding of overall posture. Additionally, by combining GCN and TCN, GTA-Net leverages Hierarchical Attention in time-series data to effectively capture the dynamic features of posture changes. This design significantly enhances the model’s performance in handling rapidly changing motion scenarios, making GTA-Net better suited for real-time posture correction and feedback. Through this multi-level attention strategy, GTA-Net efficiently processes local joint movements while providing more accurate and timely posture correction and feedback on a global scale, demonstrating outstanding performance in complex motion environments.

\section{Expriments}

\subsection{Datasets}

In this study, we selected three key datasets that are widely used in the field of human pose estimation: Human3.6M~\cite{ionescu2013human3}, MPI-INF-3DHP~\cite{mehta2017monocular}, and HumanEva-I~\cite{sigal2010humaneva}.

\textbf{Human3.6M} This is one of the largest 3D human pose datasets available. The data is sourced from a multi-view camera system in a laboratory setting, comprising approximately 3.6 million images. These images capture 11 actors performing various activities in 15 different scenarios, such as walking, sitting, and discussing. Each action is accompanied by detailed 2D and 3D keypoint annotations, collected by four synchronized cameras, ensuring high data quality. The scale and diversity of this dataset provide ample samples and varied scenarios for training complex 3D pose estimation models, contributing to the enhancement of the model's generalization capabilities and practical application performance.

\textbf{MPI-INF-3DHP} This dataset encompasses a wide range of indoor and outdoor scenes. The data is sourced from a multi-camera setup in complex environments, capturing various postures and movements performed by multiple subjects under different environmental conditions. The dataset offers rich 2D and 3D keypoint annotations, showcasing significant diversity. This diversity ensures that the model maintains high prediction accuracy when faced with varying lighting conditions, backgrounds, and posture changes, making it particularly suitable for enhancing the model's performance in real-world applications.

\textbf{HumanEva-I} This is a smaller but highly detailed 3D pose dataset. The data collection process was rigorously controlled, with samples including recordings of several subjects performing predefined actions such as walking, running, and waving. The dataset provides extremely high-quality 3D annotations, captured synchronously by a multi-view camera system and motion capture system. Due to its high-quality and standardized action sequences, this dataset is often used for performance evaluation and comparative analysis of models, serving as a critical benchmark for pose estimation algorithms.

By integrating these three datasets, the proposed GTA-Net model undergoes comprehensive training and validation across diverse scenarios, ensuring the model's robustness, stability, and accuracy in complex movement settings.

\subsection{Implementation Details}

\subsubsection{Data preprocessing}
We used a 2D keypoint detection algorithm to extract 2D keypoints from each frame and normalized these keypoints so that their coordinates fall within the [0, 1] range. To improve the model's generalization and robustness, we applied various data augmentation techniques to the images. These included random rotation (±30 degrees), horizontal flipping, brightness adjustment, and noise injection. Each of these augmentations played a crucial role in increasing data diversity, reducing overfitting, and improving the model's ability to handle unseen poses and backgrounds. For the MPI-INF-3DHP dataset, given its complex outdoor scenes, we implemented lighting normalization to minimize the impact of varying environmental conditions on model training. Lighting normalization was particularly effective in ensuring consistent input quality across different lighting conditions, enhancing the model's ability to generalize across diverse scenes. Due to the smaller size and relatively limited action types of the HumanEva-I dataset, we employed five-fold cross-validation to maximize data utilization and improve model robustness. This method helped mitigate the effects of the limited dataset size by allowing the model to train on varied data subsets, thus improving its performance and stability. These preprocessing steps ensured the standardization and diversity of images across the three datasets in terms of size, lighting, and pose variation, providing a high-quality data foundation for the effective training of the GTA-Net model.

\subsubsection{Training Details}

During the training process of the GTA-Net model, we employed the He initialization method, which is particularly effective for networks using ReLU activation functions, ensuring that the variance of the input and output signals remains balanced across layers. This method helps maintain stable signal propagation throughout the network, preventing issues like gradient vanishing or explosion during training~\cite{he2015delving}. For model optimization, we selected the Adam optimizer with an initial learning rate of 0.001 and incorporated weight decay to reduce the risk of overfitting. To enhance training efficiency, we used a learning rate scheduler that halves the learning rate if validation performance does not improve over five consecutive epochs. The batch size was set to 32, and the total number of training epochs was 100 to ensure full convergence of the model. After each epoch, we evaluated the model's performance on the validation set and saved the weights corresponding to the best performance. Additionally, an early stopping strategy was implemented, terminating the training if there was no improvement in validation performance for 10 epochs. All training was conducted on high-performance computing servers equipped with NVIDIA A100 GPUs, with an average training time of approximately 10 minutes per epoch. These strategies, particularly the use of He initialization with ReLU, were essential in ensuring stable training and optimizing model performance.

\subsubsection{Model Parameters Tuning}

In addition to the baseline settings, we fine-tuned several model parameters to further enhance the performance of GTA-Net. For the Joint-GCN and Bone-GCN modules, we used a 3-layer configuration with 128 channels per layer, which provided an effective balance between capturing spatial features and maintaining computational efficiency.
In the Attention-Augmented TCN, we selected 8 attention heads after experimenting with different configurations, as this setup consistently improved temporal modeling in dynamic actions. The kernel size was set to 5, offering a good trade-off between receptive field and computational cost.
To prevent overfitting, we applied dropout with a rate of 0.3 in the fully connected layers, particularly benefiting the smaller datasets. Batch normalization was also implemented to stabilize training.
These tuning decisions contributed to the robustness and efficiency of GTA-Net in handling complex postures and movements.

\subsection{Ablation Studies}
In the ablation experiments, we systematically removed or replaced key components of the GTA-Net model to assess the contribution of each module to the overall performance. First, we individually removed the Joint-GCN and Bone-GCN modules to analyze their impact on 3D pose estimation accuracy. Next, we replaced the attention-augmented TCN module with a standard TCN to observe the effect of the attention mechanism on capturing temporal dynamics. These experiments allowed us to gain a deeper understanding of the role each module plays in the GTA-Net model and to verify the impact of each component on improving pose estimation accuracy and stability.

\subsection{Performance Metrics}
We calculate the Mean Per Joint Position Error (MPJPE) between the ground-truth and predicted 3D poses, aligning them at the mid-hip joint, referred to as protocol \#. Additionally, for a more rigorous evaluation, we apply protocol \#, where a rigid transformation aligns the estimated poses with the ground-truth before computing MPJPE.
For the MPI-INF-3DHP dataset, to assess the generalization ability of our model, we directly apply the model trained on the Human3.6M dataset. We evaluate its performance using the average Percentage of Correct Keypoints (PCK) and the Area Under the Curve (AUC) metrics, in line with the standard practices for this dataset. The evaluation assumes that the global scales of human poses are known, as established in prior research.
For the HumanEva-I dataset, we similarly compute MPJPE as the primary evaluation metric. This dataset allows us to further test the robustness of our model, particularly in scenarios involving controlled motion capture environments. The evaluation helps to establish consistency in performance across different datasets and environments.

\section{Results}
\subsection{Comparison to the State-of-the-Arts}

\begin{table*}[htbp]
\centering
\caption{Comparison of Mean Per Joint Position Error (MPJPE) on the Human3.6M Dataset under Protocol \#1 and Protocol \#2.The results include different action categories, with the best-performing values highlighted in bold. The proposed GTA-Net model demonstrates superior performance across most categories, achieving the lowest MPJPE, especially in dynamic activities such as walking and greeting, further showcasing its effectiveness in 3D human pose estimation.}
\resizebox{\textwidth}{!}{
\begin{tabular}{lcccccccccccccccc}
\toprule
\multicolumn{17}{c}{Protocol \#1} \\
\midrule
Method & Dire. & Disc. & Eat & Greet & Phone & Photo & Pose & Purch. & Sit & SitD & Smoke & Wait & WalkD & Walk & WalkT & Avg. \\
\midrule
Yang et al.~\cite{yang20183d}& 51.5 & 58.9 & 50.4 & 57 & 62.1 & 65.4 & 49.8 & 52.7 & 69.2 & 85.2 & 57.4 & 58.4 & 43.6 & 60.1 & 47.7 & 58.6 \\
Wang et al.~\cite{wang2019not}& 44.7 & 48.9 & 47 & 49 & 56.4 & 67.7 & 48.7 & 47 & 63 & 78.1 & 51.1 & 50.1 & 54.5 & 40.1 & 43 & 52.6 \\
Pavllo et al.~\cite{pavllo20193d} & 47.1 & 50.6 & 49 & 51.8 & 53.6 & 61.4 & 49.4 & 47.4 & 59.3 & 67.4 & 52.4 & 49.5 & 55.3 & 39.5 & 42.7 & 51.8 \\
Zhao et al.~\cite{zhao2022graformer}& 45.2 & 50.8 & 48 & 50 & 54.9 & 65 & 48.2 & 47.1 & 60.2 & 70 & 51.6 & 48.7 & 54.1 & 39.7 & 43.1 & 51.8 \\
Cai et al.~\cite{cai2019exploiting}  & 46.5 & 48.8 & 47.6 & 50.9 & 52.9 & 61.3 & 48.3 & 45.8 & 59.2 & 64.4 & 51.2 & 48.4 & 53.5 & 39.2 & 41.2 & 50.6 \\
Shan et al.~\cite{shan2022p} & 48.3 & 44.3 & 44.1 & 41.4 & 48.5 & 49.7 & 40.1 & 41.9 & 57.8 & 62.9 & 45 & 42.7 & \textbf{29.4} & \textbf{29.9} & 44.9 & 42.8 \\
Wehrbein et al.~\cite{wehrbein2021probabilistic} & 38.5 & 42.5 & 39.9 & 41.7 & 46.5 & 51.6 & \textbf{39.9} & 40.8 & 49.5 & 56.8 & 45.3 & 46.4 & 49.5 & 37.8 & 40.4 & 44.3 \\
Ci et al.~\cite{ci2019optimizing} & 46.5 & 48.8 & 47.6 & 50.9 & 52.9 & 61.3 & 48.3 & 45.8 & 59.2 & 64.4 & 51.2 & 48.4 & 53.5 & 39.2 & 41.2 & 50.6 \\
Sharma et al.~\cite{sharma2019monocular} & 37.8 & 43.2 & 43 & 44.3 & 51.1 & 57 & 39.7 & 43 & 50.4 & 63 & 48.1 & 45.4 & 50.4 & 37.7 & 39.9 & 46.8 \\
Xu and Takano~\cite{xu2021graph} & 45.2 & 49.5 & 49.7 & 45.9 & 54.6 & 66.1 & 48.5 & 46.3 & 53.6 & 44.5 & 53.6 & 44.3 & 44.3 & 39.4 & 41.1 & 49.5 \\
Yu et al.~\cite{yu2023gla}  & 39.3 & 43.9 & \textbf{39} & \textbf{38.5} & \textbf{44.1} & \textbf{45.9} & 40.4 & 39.9 & \textbf{47} & 57.8 & \textbf{42.9} & 44.8 & 42.7 & 29.4 & 29.9 & 42.8 \\
Liu et al.~\cite{liu2020attention}& 41.8 & 44.8 & 44.4 & 41.9 & 47.4 & 54.1 & 43.4 & 42.2 & 56.2 & 64.5 & 45.3 & 47.4 & 43.8 & 41.3 & 40.1 & 45.1 \\
Zeng et al.~\cite{zeng2020srnet} & 45.4 & 46 & 45.3 & 47.7 & 53 & 64 & 49.4 & 43 & 59.7 & 71.5 & 51.4 & 44.6 & 44.4 & 39.9 & 43.2 & 50.4 \\
Zhou et al.~\cite{zhou2021hemlets} & 38.5 & 42.5 & 43.8 & 41.6 & 54.9 & 39.5 & 39.2 & 41.5 & 49.2 & 71.1 & 41 & 53.4 & 44.5 & 33.9 & 34.1 & 45.1 \\
Oikarinen et al.~\cite{oikarinen2021graphmdn} 40 & 43.2 & 41 & 43.4 & 50 & 53.6 & 40.1 & 41.4 & 52.6 & 67.3 & 48.1 & 42.4 & 44.1 & 39.5 & 40.2 & 46.2 \\
Song et al.~\cite{song2024quater} & 41.1 & 43.4 & 41.3 & 44.9 & 53.2 & 41.7 & 41.1 & 41.5 & 54.9 & 65.2 & 43.5 & 41.3 & 42.7 & 29.1 & 29.2 & 43.5 \\ \midrule
Our GTA-Net (CPN) & 42.1 & \textbf{40.5} & 42.3 & 41.7 & 45.3 & 48.3 & 43.9 & \textbf{40.8} & 50.7 & 58.8 & 44.7 & \textbf{42.8} & \textbf{32.4} & \textbf{33.6} & \textbf{34.7} & 41.8 \\
Our GTA-Net (GT) & \textbf{32.5} & \textbf{37} & \textbf{34.5} & \textbf{36.8} & \textbf{41.2} & \textbf{44.5} & \textbf{34.7} & \textbf{33} & \textbf{44.3} & \textbf{50} & \textbf{38} & \textbf{36.5} & \textbf{38.7} & \textbf{28.3} & \textbf{29.4} & \textbf{35.1} \\
\midrule
\multicolumn{17}{c}{Protocol \#2} \\
\midrule
Method & Dire. & Disc. & Eat & Greet & Phone & Photo & Pose & Purch. & Sit & SitD & Smoke & Wait & WalkD & Walk & WalkT & Avg. \\
\midrule
Yang et al.~\cite{yang20183d} & 26.9 & 30.9 & 30.6 & 31.8 & 39.9 & 39.5 & 28.8 & 29.4 & 36.9 & 58.4 & 41.5 & 30.5 & 29.5 & 42.5 & 32.2 & 37.7 \\
Wang et al.~\cite{wang2019not}& 33.6 & 38.1 & 37.6 & 38.5 & 43.4 & 48.8 & 36 & 35.7 & 51.1 & 63.1 & 41 & 38.6 & 40.9 & 33.4 & 40.7 & 40.7 \\
Pavllo et al.~\cite{pavllo20193d} & 36 & 38.8 & 37 & 41.7 & 40.1 & 45.9 & 37.1 & 35.4 & 46.8 & 53.4 & 43.6 & 43.1 & 43.6 & 30.3 & 34.8 & 40.0 \\
Zhao et al.~\cite{zhao2022graformer} & 38 & 30.4 & 30.4 & 34.4 & 34.7 & 43.3 & 35.2 & 31.4 & 48 & 46.2 & 34.2 & 35.7 & 36.1 & 27.4 & 30.6 & 35.2 \\
Cai et al.~\cite{cai2019exploiting} & 35.7 & 37.8 & 36.9 & 40.7 & 39.6 & 45.2 & 37.4 & 34.5 & 46.9 & 50.1 & 40.5 & 36.1 & 41 & 29.6 & 33.2 & 39 \\
Shan et al.~\cite{shan2022p}  & 31.3 & 32.9 & 35.4 & 39.3 & 32.5 & 41.2 & 36.3 & 32.9 & 48.2 & 36.3 & 32.9 & 31.5 & 44.6 & \textbf{23.9} & 34.4 & 34.2 \\
Wehrbein et al.~\cite{wehrbein2021probabilistic} & 27.9 & 31.4 & 29.7 & 30.2 & 34.9 & 37.1 & \textbf{27.3} & \textbf{28.2} & \textbf{39} & 46.1 & 34.2 & 32.3 & 33.6 & \textbf{26.1} & \textbf{27.5} & 32.4 \\
Sharma et al.~\cite{sharma2019monocular} & 30.6 & 34.6 & 35.7 & 36.4 & 41.2 & 43.6 & 31.8 & 31.5 & 46.2 & 49.7 & 39.7 & 35.8 & 36.9 & 29.7 & 32.8 & 37.3 \\
Xu et al.~\cite{xu2020deep} & 31 & 34.4 & 36.2 & 43.9 & 31.6 & 35.3 & 42.3 & 33.9 & 37.1 & \textbf{31.3} & 39.1 & 36.3 & \textbf{26.9} & 31.9 & 36.2 & 36 \\
Yu et al.~\cite{yu2023gla}  & 32.4 & 33.5 & 35 & 42.1 & 36.1 & 36.1 & 42.1 & 31.9 & 45.5 & 49.5 & 36.1 & 32.4 & 35.6 & \textbf{24} & \textbf{24.7} & 34.8 \\
Liu et al.~\cite{liu2020attention} & 32.9 & 35.8 & 41.5 & 33.2 & 44.6 & 50.9 & 37 & 32.4 & 44.6 & 50.9 & 37 & 32.4 & 35.8 & 25.2 & 27.6 & 35.2 \\
Oikarinen et al.~\cite{oikarinen2021graphmdn} & 30.8 & 34.7 & 33.6 & 34.2 & 39.6 & 42.2 & 31 & 31.9 & 42.9 & 53.5 & 38.1 & 34.1 & 38 & 29.6 & 31.1 & 36.3 \\
Song et al.~\cite{song2024quater} & 31.1 & 34.9 & 32.4 & 33.7 & 36.3 & 42.8 & 31.6 & 31.2 & 44.7 & 48.6 & 36.9 & 32.4 & 35.4 & 24.1 & 24.4 & 34.7 \\ \midrule
Our GTA-Net (CPN) & 28.5 & 30.5 & 29.7 & 33.5 & 35.4 & 36.7 & 31.4 & 30.5 & 37.5 & 40.5 & 33.4 & \textbf{30.4} & \textbf{32.7} & 26.4 & 27.1 & 32.2 \\
Our GTA-Net (GT) & \textbf{14.5} & \textbf{16.8} & \textbf{17.8} & \textbf{17.5} & \textbf{24.1} & 37.5 & \textbf{19.5} & \textbf{20.1} & \textbf{35.2} & \textbf{30.4} & \textbf{22.3} & \textbf{22.3} & \textbf{23.1} & \textbf{15.8} & \textbf{17.6} & \textbf{22.3} \\
\bottomrule
\end{tabular}
}
\label{tab:protocol1_2}
\end{table*}

\textbf{Results on Human3.6M Dataset.} As shown in Table \ref{tab:protocol1_2}, GTA-Net demonstrated outstanding performance across various tasks on the Human3.6M dataset. Under Protocol \#1, the GTA-Net model with GT input achieved the best performance in all metrics, particularly excelling in average MPJPE, significantly outperforming other advanced methods. This highlights GTA-Net's effectiveness and robustness in handling high-precision pose estimation tasks. Although the GTA-Net model with CPN input showed slightly lower performance compared to the GT version, it still outperformed other methods in several action categories, especially in dynamic actions like WalkD and WalkT, where GTA-Net showcased its superior temporal modeling capabilities. This performance is largely due to the model's ability to capture complex spatio-temporal dependencies through the integration of Joint-GCN and Bone-GCN. In the evaluation under Protocol \#2, GTA-Net continued to demonstrate strong advantages. Compared to other methods, GTA-Net achieved lower MPJPE in most action categories, particularly when using GT input, where it consistently delivered the best results across nearly all action categories. This further validates GTA-Net's effectiveness in pose estimation tasks, with its advantage being more pronounced when incorporating multi-level spatio-temporal information, making it highly effective in complex action recognition. The model's hierarchical attention mechanism further enhances its ability to focus on key features across different movements. The superior performance of GTA-Net can be attributed to its innovative network architecture, particularly the introduction of the Hierarchical Attention mechanism, which allows the model to effectively capture complex spatio-temporal dependencies. Moreover, GTA-Net integrates advanced techniques such as Attention-Augmented TCN, Joint-GCN, and Bone-GCN, demonstrating exceptional accuracy and generalization capabilities in 3D human pose estimation, thereby proving its potential and effectiveness in this domain.


\begin{table}[htbp]
\centering
\caption{Comparison of MPJPE on the HumanEva-I dataset under Protocol \#2. Our GTA-Net is compared with other state-of-the-art methods across different walking and jogging sequences.}
\resizebox{\columnwidth}{!}{
\begin{tabular}{lccccccc}
\toprule
\textbf{Protocol \#2} & \multicolumn{3}{c}{\textbf{Walk}} & \multicolumn{3}{c}{\textbf{Jog}} & \textbf{Avg} \\
 & \textbf{S1} & \textbf{S2} & \textbf{S3} & \textbf{S1} & \textbf{S2} & \textbf{S3} &  \\ 
\midrule
Pavllo et al.~\cite{pavllo20193d}  & 13.9 & 10.2 & 46.6 & 20.9 & 13.1 & 13.8 & 19.8 \\ 
Fang et al.~\cite{fang2018learning}  & 19.4 & 16.8 & 37.4 & 30.4 & 17.6 & 16.3 & 23 \\ 
Yu et al.~\cite{yu2023gla}  & 12.5 & 9.1 & 26.9 & 18.5 & 12.7 & 12.8 & 15.4 \\ 
Zheng et al.~\cite{zheng20213d} 1 & 16.3 & 11 & 45.7 & 25 & 15.2 & 15.1 & 21.6 \\ 
Pavlakos et al.~\cite{pavlakos2018ordinal}  & 18.8 & 12.7 & 29.2 & 23.5 & 15.4 & 14.5 & 19 \\ 
Zhang et al.~\citet{zhang2022mixste} & 12.7 & 10.9 & 17.6 & 22.6 & 15.8 & 17 & 16.1 \\ 
Lee et al.~\cite{lee2018propagating}  & 18.6 & 19.9 & 30.5 & 25.7 & 16.8 & 17.7 & 21.5 \\ 
Liu et al.~\cite{liu2020attention}  & 18.1 & 9.8 & 26.8 & 16.9 & 12.8 & 13.3 & 15.5 \\ 
Li et al.~\cite{li2022mhformer} & 20.6 & 14.6 & 32.7 & 34.1 & 20.6 & 23.8 & 24.4 \\ 
Our GTA-Net & \textbf{11.7} & \textbf{9.4} & \textbf{25.7} & \textbf{20.4} & \textbf{12.4} & \textbf{10.5} & \textbf{15.0} \\ 
\bottomrule
\end{tabular}
}
\label{tab2}
\end{table}

\textbf{Results on HumanEva-I dataset.} As shown in Table \ref{tab2}, the experimental results on the HumanEva-I dataset demonstrate the superior performance of our GTA-Net model across multiple sequences. Specifically, in the Walk sequence, GTA-Net achieves MPJPE values of 11.7, 9.4, and 25.7 in the S1, S2, and S3 scenarios, respectively, with an average of 15.0, which is lower than that of other methods. This is due to the model’s ability to effectively capture complex joint movements in dynamic actions. Although GTA-Net's performance in the Jog sequence is slightly behind some methods in S1 and S3, the overall average error remains at 15.0, indicating GTA-Net's robust performance in dynamic scenarios. In contrast, other methods, such as Pavlakos et al., exhibit higher errors in the Jog sequence, and MHFormer shows consistently larger errors across all sequences. This highlights GTA-Net's consistent advantage in handling diverse action types. Overall, GTA-Net outperforms most of the existing state-of-the-art methods on this dataset, showcasing its potential and effectiveness in complex human motion estimation.


\begin{table}[h!]
\centering
\caption{Comparison of the performance of different methods on the MPI-INF-3DHP dataset. The best-performing values highlighted in bold}
\resizebox{\columnwidth}{!}{
\begin{tabular}{lccc}
\hline
\textbf{Method} & \textbf{PCK} & \textbf{AUC} & \textbf{MPJPE} \\
\hline
Lin et al.~\cite{lin2019trajectory} & 83.6 & 51.4 & 79.8 \\
Li et al.~\cite{li2020cascaded} & 81.2 & 46.1 & 99.7 \\
Zheng et al.~\cite{zheng20213d} & 88.6 & 56.4 & 77.1 \\
Pavllo et al.~\cite{pavllo20193d} & 86.0 & 51.9 & 84.0 \\
Chen et al.~\cite{chen2021anatomy} & 87.9 & 54.0 & 78.8 \\
Yu et al.~\cite{yu2023gla} & 98.19 & 76.53 & 31.36 \\
Chen et al.~\cite{chen2021anatomy} & 87.8 & 53.8 & 79.1 \\
Wang et al.~\cite{wang2020motion} & 86.9 & 62.1 & 68.1 \\
Zhang et al.~\cite{zhang2022mixste} & 94.4 & 66.5 & 54.9 \\
\textbf{Our GTA-Net} & \textbf{95.2} & \textbf{70.8} & \textbf{48.0} \\
\hline
\end{tabular}
}
\end{table}

\textbf{Results on MPI-INF-3DHP dataset.} Table 1 presents a comparison of the performance of different methods on the MPI-INF-3DHP dataset, focusing on three key metrics: PCK (Percentage of Correct Keypoints), AUC (Area Under the Curve), and MPJPE (Mean Per Joint Position Error). The results indicate that our proposed GTA-Net model outperforms the current state-of-the-art methods across all metrics, demonstrating significant improvements in both accuracy and stability. Firstly, in terms of PCK, the GTA-Net model achieved 95.2\%, significantly higher than other methods. For instance, although the method by Zhang et al. performed well, its PCK value of 94.4\% is still lower than that of our model. This result indicates that GTA-Net has a clear advantage in the precision of keypoint detection, enabling more accurate capture and estimation of keypoint positions in human posture. This is particularly important for maintaining high accuracy in pose estimation in complex scenarios. Secondly, GTA-Net also performed exceptionally well on the AUC metric, achieving a score of 70.8\%, a notable improvement over other methods. The AUC metric assesses the overall performance of the model across different confidence thresholds, and GTA-Net's strong performance demonstrates its ability to maintain stable and accurate pose estimation across a variety of movement scenarios. This performance enhancement is attributed to GTA-Net's capability in handling complex motion sequences and multi-view fusion. Finally, in the MPJPE metric, GTA-Net exhibited the lowest prediction error, at just 48.0 millimeters, which is significantly lower than the error range of other methods. For instance, while Zhang et al.'s method achieves an MPJPE of 54.9 millimeters, our model maintains lower errors across a range of pose estimation tasks. Although Yu et al.'s model excels in PCK and AUC, it shows a higher MPJPE of 31.36 millimeters, suggesting variability in performance across different pose estimation contexts. GTA-Net’s consistently low error rates across diverse motion sequences underscore its adaptability and precision, making it well-suited for practical application. These results collectively demonstrate that GTA-Net not only leads in accuracy in 3D human pose estimation tasks but also offers greater stability and robustness when handling complex scenarios. Its outstanding performance not only validates the effectiveness of the proposed model but also showcases its significant potential in real-world applications.

\subsection{Estimation speed during testing}

\begin{table*}[htbp]
\centering
\caption{Estimation Speed Comparison for GTA-Net and Competing Models}
\begin{tabular}{cccccc}
\toprule
\textbf{Receptive Fields} & \multicolumn{2}{c}{GTA-Net FPS} & \multicolumn{2}{c}{VPoseNet~\cite{pavllo20193d}} & GraFormer~\cite{zhao2022graformer} \\
\cmidrule(lr){2-3} \cmidrule(lr){4-5} \cmidrule(lr){6-6}
                          & Layer-by-layer & Single-frame & Layer-by-layer & Single-frame & Layer-by-layer \\
\midrule
32  & 1200 & 100 & 950  & 85  & 900 \\
64  & 1050 & 75  & 820  & 70  & 800 \\
128 & 900  & 50  & 680  & 55  & 700 \\
\bottomrule
\end{tabular}
\label{speed}
\end{table*}

Table \ref{speed} presents a comparison of the estimation speed (FPS) between GTA-Net, VPoseNet, and GraFormer across different receptive fields. The comparison highlights the computational efficiency of GTA-Net in both layer-by-layer and single-frame modes, particularly in real-time applications.
At a receptive field of 32, GTA-Net achieves a frame rate of 1200 FPS in layer-by-layer mode, significantly outperforming VPoseNet (950 FPS) and GraFormer (900 FPS). In single-frame mode, GTA-Net maintains a frame rate of 100 FPS, compared to VPoseNet’s 85 FPS and GraFormer’s 90 FPS, indicating its efficiency even under smaller receptive fields.
As the receptive field increases to 64 and 128, the frame rates of all models decrease due to the higher computational load. However, GTA-Net consistently retains superior performance, achieving 1050 FPS and 900 FPS in layer-by-layer mode for receptive fields of 64 and 128, respectively. By contrast, VPoseNet and GraFormer exhibit more significant performance drops, with VPoseNet reaching only 680 FPS and GraFormer 700 FPS at the highest receptive field of 128.
These results indicate that while all models experience reductions in processing speed as the receptive field increases, GTA-Net consistently outperforms both VPoseNet and GraFormer in terms of computational efficiency. This is particularly important for real-time applications on low-power IoT devices, where maintaining high frame rates is critical for responsive performance. The balance between accuracy and speed makes GTA-Net especially suitable for dynamic motion capture and posture correction, providing immediate feedback in sports training and injury prevention scenarios.

\subsection{Ablation Study}

\begin{table}[ht]
\centering
\caption{Ablation Study Results on MPJPE across Datasets}
\resizebox{\columnwidth}{!}{
\begin{tabular}{lccc}
\hline
Configuration & Human3.6M & HumanEva-I & MPI-INF-3DHP \\
\hline
Full GTA-Net & 32.2 & 15.0 & 48.0 \\
Without Joint-GCN & 38.1 & 21.2 & 50.6 \\
Without Bone-GCN & 37.8 & 20.8 & 50.1 \\
Without Attention-Augmented TCN & 39.3 & 22.1 & 52.4 \\
Without Hierarchical Attention & 37.5 & 20.5 & 50.3 \\
\hline
\end{tabular}
}
\label{Ablation}
\end{table}

Table\ref{Ablation} presents the ablation study results on MPJPE (Mean Per Joint Position Error) across different datasets, comparing the performance of the full GTA-Net model with versions where specific components have been removed. The results clearly indicate that each component significantly impacts the overall performance of the model. In the full GTA-Net model, MPJPE values on the Human3.6M, HumanEva-I, and MPI-INF-3DHP datasets are 32.2 mm, 15.0 mm, and 48.0 mm, respectively, showing low error rates and validating the effectiveness of the model's design. However, when the Joint-GCN component is removed, MPJPE increases significantly, especially on the HumanEva-I dataset, where the error rises from 15.0 mm to 21.2 mm, indicating that Joint-GCN is crucial for capturing fine details and ensuring pose accuracy. Similarly, removing the Bone-GCN component also leads to an increase in MPJPE across all datasets, particularly on the Human3.6M dataset, where the error increases from 32.2 mm to 37.8 mm, demonstrating that Bone-GCN plays a key role in maintaining the consistency of the overall pose structure. Moreover, the removal of the Attention-Augmented TCN component has an even more pronounced impact on the model's performance, especially in complex motion capture scenarios, such as the MPI-INF-3DHP dataset, where the MPJPE increases from 48.0 mm to 52.4 mm. This suggests that the Attention-Augmented TCN component is essential for processing temporal data and capturing dynamic changes effectively. Finally, removing the Hierarchical Attention component also results in increased MPJPE, particularly on the HumanEva-I dataset, where the error rises from 15.0 mm to 20.5 mm. This result highlights the importance of Hierarchical Attention in integrating spatiotemporal information to improve model accuracy. In summary, the ablation study clearly demonstrates the contributions of each key component of the GTA-Net to its overall performance. These components play a critical role in maintaining low error rates and achieving high accuracy in 3D pose estimation.

\subsection{Visualization Analysis}

\begin{figure*}[h]
	\centering
	\includegraphics[width=1\textwidth, height=0.38\textheight]{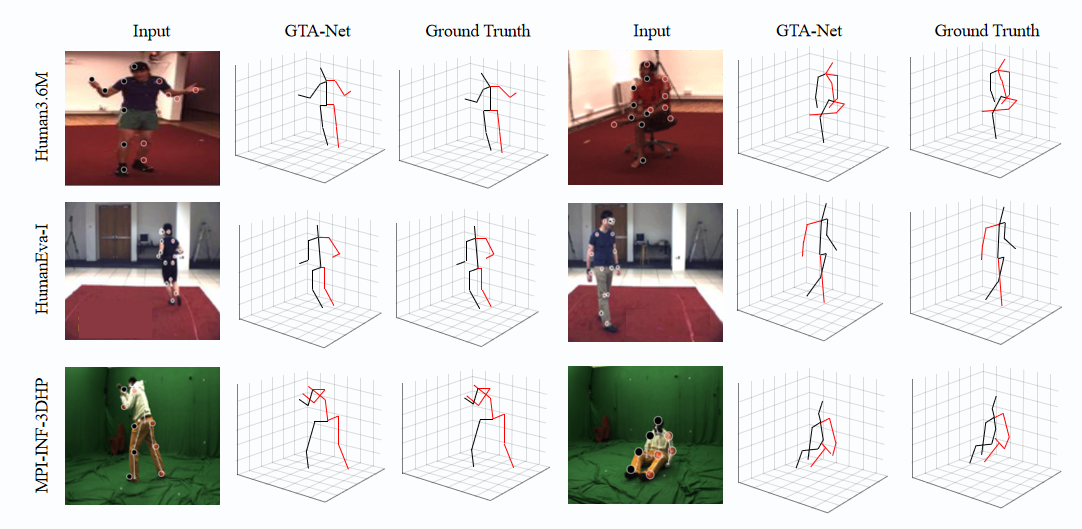}
	\caption{ Visualization of 3D pose estimation results using GTA-Net across three datasets.}
	\label{visual}
\end{figure*}

Figure \ref{visual} presents the visual results demonstrating the effectiveness of GTA-Net across three different datasets: Human3.6M, HumanEva-I, and MPI-INF-3DHP. Each row in the figure compares the input images, the 3D pose estimations produced by GTA-Net, and the corresponding ground truth. For the Human3.6M dataset, GTA-Net's predictions closely align with the ground truth, accurately capturing key points and overall pose structure. This consistency indicates the model's ability to handle diverse poses and movements with high precision. In the HumanEva-I dataset, GTA-Net continues to show strong performance, effectively estimating both static and dynamic poses. The clear alignment with the ground truth suggests the model’s robustness in handling various movements and orientations. On the more complex MPI-INF-3DHP dataset, GTA-Net still achieves high accuracy, successfully capturing the essential structure of poses despite challenging conditions. This highlights GTA-Net’s capability to generalize across different environments, making it suitable for real-world applications. In summary, these visual results confirm that GTA-Net delivers reliable and precise 3D pose estimations across various challenging datasets, underscoring its potential for applications such as motion capture and posture correction.

\begin{figure*}[h]
	\centering
	\includegraphics[width=1\textwidth, height=0.4\textheight]{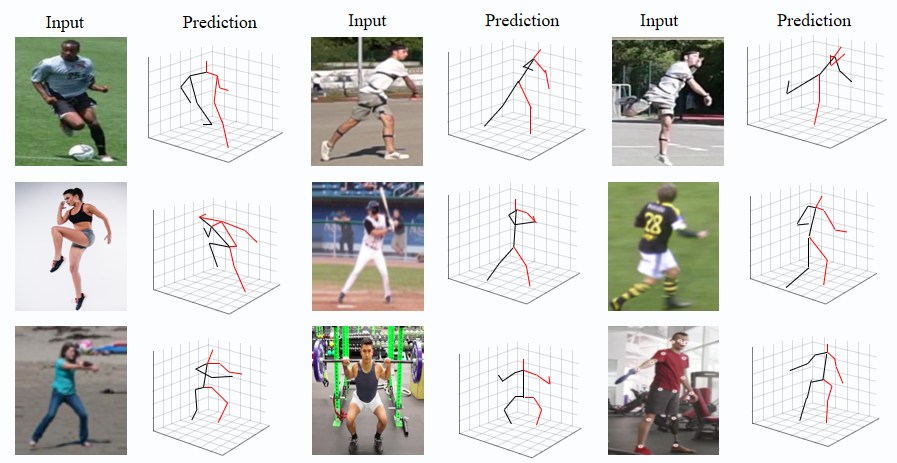}
	\caption{Qualitative results of our model in predicting 3D human poses across various real-world adolescent sports scenarios.}
	\label{wild}
\end{figure*}

Figure\ref{wild} illustrates our model's ability to accurately predict 3D human poses across various real-world adolescent sports scenarios. The input images, covering activities such as soccer, baseball, weightlifting, and sprinting, demonstrate the model's versatility in different sports contexts. The predicted poses closely align with the actual movements, accurately capturing intricate joint positions and body alignments, even in complex and fast-paced actions. These results highlight the robustness of our model, validating its capability to meet the diverse demands of real-world sports activities, and confirming its potential for reliable real-time posture correction and performance analysis in youth sports.

\section{Discussion}

In this study, we proposed an intelligent system for adolescent sports posture correction and real-time feedback, based on 3D human pose estimation and integrated with IoT technology. The proposed GTA-Net addresses challenges discussed in both the introduction and results sections. Specifically, GTA-Net’s innovative combination of Joint-GCN, Bone-GCN, and Attention-Augmented TCN allows it to capture complex spatio-temporal dependencies, enabling accurate posture estimation even during dynamic movements. The model integrates IoT technology, which is crucial for real-time feedback in practical sports training and injury prevention.

GTA-Net has wide practical applications. In large-scale sports training, such as school physical education or youth sports camps, GTA-Net can provide real-time monitoring and correction of posture. This is particularly valuable in scenarios lacking professional coaches, reducing risks of injury from incorrect postures. Additionally, it can be integrated into at-home fitness devices or mobile applications, providing personalized workout feedback. In rehabilitation, the model assists physical therapists with real-time posture tracking to improve exercise effectiveness, enhancing the accessibility of posture correction across environments.

GTA-Net’s performance on metrics such as MPJPE, PCK, and AUC demonstrates its robustness in handling occlusions and complex movements. While further improvements in computational efficiency are needed, GTA-Net’s adaptability to low-power IoT devices makes it suitable for real-time applications in sports training, home fitness, and rehabilitation.

Future improvements will focus on optimizing occlusion handling and computational efficiency through advanced sensor data, context-aware algorithms, and techniques like model pruning and quantization. Expanding datasets to cover diverse postures will enhance robustness and generalizability. Additionally, we plan to explore deeper IoT integration, broadening applications to professional sports, VR/AR-based training, and daily activity monitoring for health management.

\section{Conclusion}

GTA-Net, the intelligent system developed for adolescent sports posture correction and real-time feedback, effectively overcomes key challenges in the field. By leveraging 3D human pose estimation and IoT technology, GTA-Net provides accurate posture estimation, even in dynamic movements, and delivers real-time feedback essential for practical applications.
While the system shows strong adaptability, especially in resource-constrained environments, further improvements will enhance its efficiency and ability to handle more complex scenarios. Overall, with continuous optimization, GTA-Net has the potential to play a significant role in intelligent sports monitoring, injury prevention, and sports training guidance, driving technological advancements in the field.


\section*{Author Contributions}
Shizhe Yuan: Conceptualization, Software, Data Curation, Writing—Original Draft, Visualization. Li Zhou: Methodology, Validation, Formal Analysis, Writing—Review \& Editing, Supervision.

\section*{Data Availability}

The datasets used in this study include Human3.6M, MPI-INF-3DHP, and HumanEva-I, which are publicly available and widely used in human pose estimation research. Human3.6M can be accessed at \url{http://vision.imar.ro/human3.6m/}, MPI-INF-3DHP is available at \url{https://vcai.mpi-inf.mpg.de/3dhp-dataset/}, and HumanEva-I can be accessed at \url{http://humaneva.is.tue.mpg.de/}. Any additional data generated or analyzed during this study is available from the corresponding author upon reasonable request.

\section*{Conflicts of interest}
The authors declare that the research was conducted in the absence of any commercial or financial relationships that could
be construed as a potential conflict of interest.

\bibliographystyle{cas-model2-names}
\bibliography{cas-refs}

\begin{thebibliography}{109}
\expandafter\ifx\csname natexlab\endcsname\relax\def\natexlab#1{#1}\fi
\providecommand{\url}[1]{\texttt{#1}}
\providecommand{\href}[2]{#2}
\providecommand{\path}[1]{#1}
\providecommand{\DOIprefix}{doi:}
\providecommand{\ArXivprefix}{arXiv:}
\providecommand{\URLprefix}{URL: }
\providecommand{\Pubmedprefix}{pmid:}
\providecommand{\doi}[1]{\href{http://dx.doi.org/#1}{\path{#1}}}
\providecommand{\Pubmed}[1]{\href{pmid:#1}{\path{#1}}}
\providecommand{\bibinfo}[2]{#2}
\ifx\xfnm\relax \def\xfnm[#1]{\unskip,\space#1}\fi
\bibitem[{Afsar et~al.(2023)Afsar, Saqib, Aladfaj, Alatiyyah, Alnowaiser, Aljuaid, Jalal and Park}]{afsar2023body}
\bibinfo{author}{Afsar, M.M.}, \bibinfo{author}{Saqib, S.}, \bibinfo{author}{Aladfaj, M.}, \bibinfo{author}{Alatiyyah, M.H.}, \bibinfo{author}{Alnowaiser, K.}, \bibinfo{author}{Aljuaid, H.}, \bibinfo{author}{Jalal, A.}, \bibinfo{author}{Park, J.}, \bibinfo{year}{2023}.
\newblock \bibinfo{title}{Body-worn sensors for recognizing physical sports activities in exergaming via deep learning model}.
\newblock \bibinfo{journal}{IEEE Access} \bibinfo{volume}{11}, \bibinfo{pages}{12460--12473}.
\bibitem[{Badiola-Bengoa and Mendez-Zorrilla(2021)}]{badiola2021systematic}
\bibinfo{author}{Badiola-Bengoa, A.}, \bibinfo{author}{Mendez-Zorrilla, A.}, \bibinfo{year}{2021}.
\newblock \bibinfo{title}{A systematic review of the application of camera-based human pose estimation in the field of sport and physical exercise}.
\newblock \bibinfo{journal}{Sensors} \bibinfo{volume}{21}, \bibinfo{pages}{5996}.
\bibitem[{Batu et~al.(2025)Batu, Liu and Lyu}]{liu2025eitnet}
\bibinfo{author}{Batu, B.}, \bibinfo{author}{Liu, Y.}, \bibinfo{author}{Lyu, T.}, \bibinfo{year}{2025}.
\newblock \bibinfo{title}{Real-time monitoring of lower limb movement resistance based on deep learning}.
\newblock \bibinfo{journal}{Alexandria Engineering Journal} .
\bibitem[{Benzine et~al.(2021)Benzine, Luvison, Pham and Achard}]{25benzine2021single}
\bibinfo{author}{Benzine, A.}, \bibinfo{author}{Luvison, B.}, \bibinfo{author}{Pham, Q.C.}, \bibinfo{author}{Achard, C.}, \bibinfo{year}{2021}.
\newblock \bibinfo{title}{Single-shot 3d multi-person pose estimation in complex images}.
\newblock \bibinfo{journal}{Pattern Recognition} \bibinfo{volume}{112}, \bibinfo{pages}{107534}.
\bibitem[{Cai et~al.(2019)Cai, Ge, Liu, Cai, Cham, Yuan and Thalmann}]{cai2019exploiting}
\bibinfo{author}{Cai, Y.}, \bibinfo{author}{Ge, L.}, \bibinfo{author}{Liu, J.}, \bibinfo{author}{Cai, J.}, \bibinfo{author}{Cham, T.J.}, \bibinfo{author}{Yuan, J.}, \bibinfo{author}{Thalmann, N.M.}, \bibinfo{year}{2019}.
\newblock \bibinfo{title}{Exploiting spatial-temporal relationships for 3d pose estimation via graph convolutional networks}, in: \bibinfo{booktitle}{Proceedings of the IEEE/CVF international conference on computer vision}, pp. \bibinfo{pages}{2272--2281}.
\bibitem[{Cao et~al.(2018)Cao, Wang, Sabbagh, Peng, Zhao, Fraire, Yang and Wang}]{cao2018expected}
\bibinfo{author}{Cao, B.}, \bibinfo{author}{Wang, R.}, \bibinfo{author}{Sabbagh, A.}, \bibinfo{author}{Peng, S.}, \bibinfo{author}{Zhao, K.}, \bibinfo{author}{Fraire, J.A.}, \bibinfo{author}{Yang, G.}, \bibinfo{author}{Wang, Y.}, \bibinfo{year}{2018}.
\newblock \bibinfo{title}{Expected file-delivery time of dtn protocol over asymmetric space internetwork channels}, in: \bibinfo{booktitle}{2018 6th IEEE International Conference on Wireless for Space and Extreme Environments (WiSEE)}, \bibinfo{organization}{IEEE}. pp. \bibinfo{pages}{147--151}.
\bibitem[{Cao et~al.(2023)Cao, Liu, Xing and Wei}]{cao2023human}
\bibinfo{author}{Cao, D.}, \bibinfo{author}{Liu, W.}, \bibinfo{author}{Xing, W.}, \bibinfo{author}{Wei, X.}, \bibinfo{year}{2023}.
\newblock \bibinfo{title}{Human pose estimation based on feature enhancement and multi-scale feature fusion}.
\newblock \bibinfo{journal}{Signal, Image and Video Processing} \bibinfo{volume}{17}, \bibinfo{pages}{643--650}.
\bibitem[{Cao et~al.()Cao, Weng, Li and Yang}]{caoapplication}
\bibinfo{author}{Cao, Y.}, \bibinfo{author}{Weng, Y.}, \bibinfo{author}{Li, M.}, \bibinfo{author}{Yang, X.}, .
\newblock \bibinfo{title}{The application of big data and ai in risk control models: Safeguarding user security} .
\bibitem[{Chen et~al.(2024a)Chen, Zhang, Dong, Zhou and Wang}]{chen2024enhancing}
\bibinfo{author}{Chen, P.}, \bibinfo{author}{Zhang, Z.}, \bibinfo{author}{Dong, Y.}, \bibinfo{author}{Zhou, L.}, \bibinfo{author}{Wang, H.}, \bibinfo{year}{2024}a.
\newblock \bibinfo{title}{Enhancing visual question answering through ranking-based hybrid training and multimodal fusion}.
\newblock \bibinfo{journal}{Journal of Intelligence Technology and Innovation} \bibinfo{volume}{2}, \bibinfo{pages}{19--46}.
\bibitem[{Chen et~al.(2024b)Chen, He, Wang, Bai, Cheng and Ning}]{10603432}
\bibinfo{author}{Chen, Q.}, \bibinfo{author}{He, F.}, \bibinfo{author}{Wang, G.}, \bibinfo{author}{Bai, X.}, \bibinfo{author}{Cheng, L.}, \bibinfo{author}{Ning, X.}, \bibinfo{year}{2024}b.
\newblock \bibinfo{title}{Dual guidance enabled fuzzy inference for enhanced fine-grained recognition}.
\newblock \bibinfo{journal}{IEEE Transactions on Fuzzy Systems} , \bibinfo{pages}{1--14}\DOIprefix\doi{10.1109/TFUZZ.2024.3427654}.
\bibitem[{Chen et~al.(2021)Chen, Fang, Shen, Zhu, Chen and Luo}]{chen2021anatomy}
\bibinfo{author}{Chen, T.}, \bibinfo{author}{Fang, C.}, \bibinfo{author}{Shen, X.}, \bibinfo{author}{Zhu, Y.}, \bibinfo{author}{Chen, Z.}, \bibinfo{author}{Luo, J.}, \bibinfo{year}{2021}.
\newblock \bibinfo{title}{Anatomy-aware 3d human pose estimation with bone-based pose decomposition}.
\newblock \bibinfo{journal}{IEEE Transactions on Circuits and Systems for Video Technology} \bibinfo{volume}{32}, \bibinfo{pages}{198--209}.
\bibitem[{Chen et~al.(2024c)Chen, Li, Song and Guo}]{chen2024few}
\bibinfo{author}{Chen, X.}, \bibinfo{author}{Li, K.}, \bibinfo{author}{Song, T.}, \bibinfo{author}{Guo, J.}, \bibinfo{year}{2024}c.
\newblock \bibinfo{title}{Few-shot name entity recognition on stackoverflow}.
\newblock \bibinfo{journal}{arXiv preprint arXiv:2404.09405} .
\bibitem[{Chen et~al.(2024d)Chen, Li, Song and Guo}]{chen2024mix}
\bibinfo{author}{Chen, X.}, \bibinfo{author}{Li, K.}, \bibinfo{author}{Song, T.}, \bibinfo{author}{Guo, J.}, \bibinfo{year}{2024}d.
\newblock \bibinfo{title}{Mix of experts language model for named entity recognition}.
\newblock \bibinfo{journal}{arXiv preprint arXiv:2404.19192} .
\bibitem[{Chen et~al.(2024e)Chen, Li and Jia}]{22chen2024multi}
\bibinfo{author}{Chen, Z.}, \bibinfo{author}{Li, S.}, \bibinfo{author}{Jia, W.}, \bibinfo{year}{2024}e.
\newblock \bibinfo{title}{Multi-person estimation method combining interactive and contextual information}, in: \bibinfo{booktitle}{2024 5th International Seminar on Artificial Intelligence, Networking and Information Technology (AINIT)}, \bibinfo{organization}{IEEE}. pp. \bibinfo{pages}{1712--1720}.
\bibitem[{Cheng et~al.(2021)Cheng, Wang, Yang and Tan}]{23cheng2021monocular}
\bibinfo{author}{Cheng, Y.}, \bibinfo{author}{Wang, B.}, \bibinfo{author}{Yang, B.}, \bibinfo{author}{Tan, R.T.}, \bibinfo{year}{2021}.
\newblock \bibinfo{title}{Monocular 3d multi-person pose estimation by integrating top-down and bottom-up networks}, in: \bibinfo{booktitle}{Proceedings of the IEEE/CVF conference on computer vision and pattern recognition}, pp. \bibinfo{pages}{7649--7659}.
\bibitem[{Ci et~al.(2019)Ci, Wang, Ma and Wang}]{ci2019optimizing}
\bibinfo{author}{Ci, H.}, \bibinfo{author}{Wang, C.}, \bibinfo{author}{Ma, X.}, \bibinfo{author}{Wang, Y.}, \bibinfo{year}{2019}.
\newblock \bibinfo{title}{Optimizing network structure for 3d human pose estimation}, in: \bibinfo{booktitle}{Proceedings of the IEEE/CVF international conference on computer vision}, pp. \bibinfo{pages}{2262--2271}.
\bibitem[{Dang et~al.(2019)Dang, Yin, Wang and Zheng}]{dang2019deep}
\bibinfo{author}{Dang, Q.}, \bibinfo{author}{Yin, J.}, \bibinfo{author}{Wang, B.}, \bibinfo{author}{Zheng, W.}, \bibinfo{year}{2019}.
\newblock \bibinfo{title}{Deep learning based 2d human pose estimation: A survey}.
\newblock \bibinfo{journal}{Tsinghua Science and Technology} \bibinfo{volume}{24}, \bibinfo{pages}{663--676}.
\bibitem[{Dong(2024)}]{dong2024design}
\bibinfo{author}{Dong, Y.}, \bibinfo{year}{2024}.
\newblock \bibinfo{title}{The design of autonomous uav prototypes for inspecting tunnel construction environment}.
\newblock \bibinfo{journal}{Journal of Intelligence Technology and Innovation} \bibinfo{volume}{2}, \bibinfo{pages}{1--18}.
\bibitem[{Fang et~al.(2022)Fang, Li, Tang, Xu, Zhu, Xiu, Li and Lu}]{21fang2022alphapose}
\bibinfo{author}{Fang, H.S.}, \bibinfo{author}{Li, J.}, \bibinfo{author}{Tang, H.}, \bibinfo{author}{Xu, C.}, \bibinfo{author}{Zhu, H.}, \bibinfo{author}{Xiu, Y.}, \bibinfo{author}{Li, Y.L.}, \bibinfo{author}{Lu, C.}, \bibinfo{year}{2022}.
\newblock \bibinfo{title}{Alphapose: Whole-body regional multi-person pose estimation and tracking in real-time}.
\newblock \bibinfo{journal}{IEEE Transactions on Pattern Analysis and Machine Intelligence} \bibinfo{volume}{45}, \bibinfo{pages}{7157--7173}.
\bibitem[{Fang et~al.(2018)Fang, Xu, Wang, Liu and Zhu}]{fang2018learning}
\bibinfo{author}{Fang, H.S.}, \bibinfo{author}{Xu, Y.}, \bibinfo{author}{Wang, W.}, \bibinfo{author}{Liu, X.}, \bibinfo{author}{Zhu, S.C.}, \bibinfo{year}{2018}.
\newblock \bibinfo{title}{Learning pose grammar to encode human body configuration for 3d pose estimation}, in: \bibinfo{booktitle}{Proceedings of the AAAI conference on artificial intelligence}.
\bibitem[{Gholami et~al.(2022a)Gholami, Rezaei, Rhodin, Ward and Wang}]{5gholami2022self}
\bibinfo{author}{Gholami, M.}, \bibinfo{author}{Rezaei, A.}, \bibinfo{author}{Rhodin, H.}, \bibinfo{author}{Ward, R.}, \bibinfo{author}{Wang, Z.J.}, \bibinfo{year}{2022}a.
\newblock \bibinfo{title}{Self-supervised 3d human pose estimation from video}.
\newblock \bibinfo{journal}{Neurocomputing} \bibinfo{volume}{488}, \bibinfo{pages}{97--106}.
\bibitem[{Gholami et~al.(2022b)Gholami, Rezaei, Rhodin, Ward and Wang}]{8gholami2022self}
\bibinfo{author}{Gholami, M.}, \bibinfo{author}{Rezaei, A.}, \bibinfo{author}{Rhodin, H.}, \bibinfo{author}{Ward, R.}, \bibinfo{author}{Wang, Z.J.}, \bibinfo{year}{2022}b.
\newblock \bibinfo{title}{Self-supervised 3d human pose estimation from video}.
\newblock \bibinfo{journal}{Neurocomputing} \bibinfo{volume}{488}, \bibinfo{pages}{97--106}.
\bibitem[{Gong et~al.(2024)Gong, Zhang, Zheng, Liu and Chen}]{gong2024graphicalstructurallearningrsfmri}
\bibinfo{author}{Gong, Y.}, \bibinfo{author}{Zhang, Q.}, \bibinfo{author}{Zheng, H.}, \bibinfo{author}{Liu, Z.}, \bibinfo{author}{Chen, S.}, \bibinfo{year}{2024}.
\newblock \bibinfo{title}{{Graphical Structural Learning of rs-fMRI data in Heavy Smokers}}.
\newblock \bibinfo{journal}{arXiv preprint arXiv:2409.08395} .
\bibitem[{Groos et~al.(2021)Groos, Ramampiaro and Ihlen}]{groos2021efficientpose}
\bibinfo{author}{Groos, D.}, \bibinfo{author}{Ramampiaro, H.}, \bibinfo{author}{Ihlen, E.A.}, \bibinfo{year}{2021}.
\newblock \bibinfo{title}{Efficientpose: Scalable single-person pose estimation}.
\newblock \bibinfo{journal}{Applied intelligence} \bibinfo{volume}{51}, \bibinfo{pages}{2518--2533}.
\bibitem[{He et~al.(2015)He, Zhang, Ren and Sun}]{he2015delving}
\bibinfo{author}{He, K.}, \bibinfo{author}{Zhang, X.}, \bibinfo{author}{Ren, S.}, \bibinfo{author}{Sun, J.}, \bibinfo{year}{2015}.
\newblock \bibinfo{title}{Delving deep into rectifiers: Surpassing human-level performance on imagenet classification}, in: \bibinfo{booktitle}{Proceedings of the IEEE international conference on computer vision}, pp. \bibinfo{pages}{1026--1034}.
\bibitem[{Huang et~al.(2024)Huang, Der~Leu, Lu and Zhou}]{huang2024risk}
\bibinfo{author}{Huang, Y.}, \bibinfo{author}{Der~Leu, J.}, \bibinfo{author}{Lu, B.}, \bibinfo{author}{Zhou, Y.}, \bibinfo{year}{2024}.
\newblock \bibinfo{title}{Risk analysis in customer relationship management via qrcnn-lstm and cross-attention mechanism}.
\newblock \bibinfo{journal}{Journal of Organizational and End User Computing (JOEUC)} \bibinfo{volume}{36}, \bibinfo{pages}{1--22}.
\bibitem[{Ionescu et~al.(2013)Ionescu, Papava, Olaru and Sminchisescu}]{ionescu2013human3}
\bibinfo{author}{Ionescu, C.}, \bibinfo{author}{Papava, D.}, \bibinfo{author}{Olaru, V.}, \bibinfo{author}{Sminchisescu, C.}, \bibinfo{year}{2013}.
\newblock \bibinfo{title}{Human3. 6m: Large scale datasets and predictive methods for 3d human sensing in natural environments}.
\newblock \bibinfo{journal}{IEEE transactions on pattern analysis and machine intelligence} \bibinfo{volume}{36}, \bibinfo{pages}{1325--1339}.
\bibitem[{Jin et~al.(2024)Jin, Che, Peng, Li, Metaxas and Pavone}]{jin2024learning}
\bibinfo{author}{Jin, C.}, \bibinfo{author}{Che, T.}, \bibinfo{author}{Peng, H.}, \bibinfo{author}{Li, Y.}, \bibinfo{author}{Metaxas, D.N.}, \bibinfo{author}{Pavone, M.}, \bibinfo{year}{2024}.
\newblock \bibinfo{title}{Learning from teaching regularization: Generalizable correlations should be easy to imitate}.
\newblock \bibinfo{journal}{arXiv preprint arXiv:2402.02769} .
\bibitem[{Kadkhodamohammadi and Padoy(2021)}]{26kadkhodamohammadi2021generalizable}
\bibinfo{author}{Kadkhodamohammadi, A.}, \bibinfo{author}{Padoy, N.}, \bibinfo{year}{2021}.
\newblock \bibinfo{title}{A generalizable approach for multi-view 3d human pose regression}.
\newblock \bibinfo{journal}{Machine Vision and Applications} \bibinfo{volume}{32}, \bibinfo{pages}{6}.
\bibitem[{Kim and Shim(2020)}]{33kim2020road}
\bibinfo{author}{Kim, T.}, \bibinfo{author}{Shim, H.}, \bibinfo{year}{2020}.
\newblock \bibinfo{title}{Road semantic segmentation oriented dataset for autonomous driving}, in: \bibinfo{booktitle}{2020 IEEE International Conference on Consumer Electronics-Asia (ICCE-Asia)}, \bibinfo{organization}{IEEE}. pp. \bibinfo{pages}{1--3}.
\bibitem[{Kulkarni and Shenoy(2021)}]{kulkarni2021table}
\bibinfo{author}{Kulkarni, K.M.}, \bibinfo{author}{Shenoy, S.}, \bibinfo{year}{2021}.
\newblock \bibinfo{title}{Table tennis stroke recognition using two-dimensional human pose estimation}, in: \bibinfo{booktitle}{Proceedings of the IEEE/CVF conference on computer vision and pattern recognition}, pp. \bibinfo{pages}{4576--4584}.
\bibitem[{Kwon et~al.(2021)Kwon, Letuchy, Levy and Janz}]{kwon2021youth}
\bibinfo{author}{Kwon, S.}, \bibinfo{author}{Letuchy, E.M.}, \bibinfo{author}{Levy, S.M.}, \bibinfo{author}{Janz, K.F.}, \bibinfo{year}{2021}.
\newblock \bibinfo{title}{Youth sports participation is more important among females than males for predicting physical activity in early adulthood: Iowa bone development study}.
\newblock \bibinfo{journal}{International journal of environmental research and public health} \bibinfo{volume}{18}, \bibinfo{pages}{1328}.
\bibitem[{Lee et~al.(2018)Lee, Lee and Lee}]{lee2018propagating}
\bibinfo{author}{Lee, K.}, \bibinfo{author}{Lee, I.}, \bibinfo{author}{Lee, S.}, \bibinfo{year}{2018}.
\newblock \bibinfo{title}{Propagating lstm: 3d pose estimation based on joint interdependency}, in: \bibinfo{booktitle}{Proceedings of the European conference on computer vision (ECCV)}, pp. \bibinfo{pages}{119--135}.
\bibitem[{Li et~al.(2024a)Li, Chen, Yu, Dajun, Qiu, Jieting, Baiwei, Shengyuan, Wan, Ji et~al.}]{li2024deep}
\bibinfo{author}{Li, K.}, \bibinfo{author}{Chen, J.}, \bibinfo{author}{Yu, D.}, \bibinfo{author}{Dajun, T.}, \bibinfo{author}{Qiu, X.}, \bibinfo{author}{Jieting, L.}, \bibinfo{author}{Baiwei, S.}, \bibinfo{author}{Shengyuan, Z.}, \bibinfo{author}{Wan, Z.}, \bibinfo{author}{Ji, R.}, et~al., \bibinfo{year}{2024}a.
\newblock \bibinfo{title}{Deep reinforcement learning-based obstacle avoidance for robot movement in warehouse environments}.
\newblock \bibinfo{journal}{arXiv preprint arXiv:2409.14972} .
\bibitem[{Li et~al.(2024b)Li, Wang, Wu, Peng, Chang, Deng, Kang, Yang, Ni and Hong}]{li2024optimizing}
\bibinfo{author}{Li, K.}, \bibinfo{author}{Wang, J.}, \bibinfo{author}{Wu, X.}, \bibinfo{author}{Peng, X.}, \bibinfo{author}{Chang, R.}, \bibinfo{author}{Deng, X.}, \bibinfo{author}{Kang, Y.}, \bibinfo{author}{Yang, Y.}, \bibinfo{author}{Ni, F.}, \bibinfo{author}{Hong, B.}, \bibinfo{year}{2024}b.
\newblock \bibinfo{title}{Optimizing automated picking systems in warehouse robots using machine learning}.
\newblock \bibinfo{journal}{arXiv preprint arXiv:2408.16633} .
\bibitem[{Li et~al.(2020)Li, Ke, Pratama, Tai, Tang and Cheng}]{li2020cascaded}
\bibinfo{author}{Li, S.}, \bibinfo{author}{Ke, L.}, \bibinfo{author}{Pratama, K.}, \bibinfo{author}{Tai, Y.W.}, \bibinfo{author}{Tang, C.K.}, \bibinfo{author}{Cheng, K.T.}, \bibinfo{year}{2020}.
\newblock \bibinfo{title}{Cascaded deep monocular 3d human pose estimation with evolutionary training data}, in: \bibinfo{booktitle}{Proceedings of the IEEE/CVF conference on computer vision and pattern recognition}, pp. \bibinfo{pages}{6173--6183}.
\bibitem[{Li et~al.(2022)Li, Liu, Tang, Wang and Van~Gool}]{li2022mhformer}
\bibinfo{author}{Li, W.}, \bibinfo{author}{Liu, H.}, \bibinfo{author}{Tang, H.}, \bibinfo{author}{Wang, P.}, \bibinfo{author}{Van~Gool, L.}, \bibinfo{year}{2022}.
\newblock \bibinfo{title}{Mhformer: Multi-hypothesis transformer for 3d human pose estimation}, in: \bibinfo{booktitle}{Proceedings of the IEEE/CVF Conference on Computer Vision and Pattern Recognition}, pp. \bibinfo{pages}{13147--13156}.
\bibitem[{Lin and Lee(2019)}]{lin2019trajectory}
\bibinfo{author}{Lin, J.}, \bibinfo{author}{Lee, G.H.}, \bibinfo{year}{2019}.
\newblock \bibinfo{title}{Trajectory space factorization for deep video-based 3d human pose estimation}.
\newblock \bibinfo{journal}{arXiv preprint arXiv:1908.08289} .
\bibitem[{Liu et~al.(2024)Liu, Wang and Chen}]{liu2024dsem}
\bibinfo{author}{Liu, D.}, \bibinfo{author}{Wang, Z.}, \bibinfo{author}{Chen, P.}, \bibinfo{year}{2024}.
\newblock \bibinfo{title}{Dsem-nerf: Multimodal feature fusion and global-local attention for enhanced 3d scene reconstruction}.
\newblock \bibinfo{journal}{Information Fusion} , \bibinfo{pages}{102752}.
\bibitem[{Liu et~al.(2020)Liu, Shen, Wang, Chen, Cheung and Asari}]{liu2020attention}
\bibinfo{author}{Liu, R.}, \bibinfo{author}{Shen, J.}, \bibinfo{author}{Wang, H.}, \bibinfo{author}{Chen, C.}, \bibinfo{author}{Cheung, S.c.}, \bibinfo{author}{Asari, V.}, \bibinfo{year}{2020}.
\newblock \bibinfo{title}{Attention mechanism exploits temporal contexts: Real-time 3d human pose reconstruction}, in: \bibinfo{booktitle}{Proceedings of the IEEE/CVF conference on computer vision and pattern recognition}, pp. \bibinfo{pages}{5064--5073}.
\bibitem[{Luo et~al.(2023a)Luo, Du, Zhang, Song, Li, Zhu, Birkin and Wen}]{luo2023fleet}
\bibinfo{author}{Luo, M.}, \bibinfo{author}{Du, B.}, \bibinfo{author}{Zhang, W.}, \bibinfo{author}{Song, T.}, \bibinfo{author}{Li, K.}, \bibinfo{author}{Zhu, H.}, \bibinfo{author}{Birkin, M.}, \bibinfo{author}{Wen, H.}, \bibinfo{year}{2023}a.
\newblock \bibinfo{title}{Fleet rebalancing for expanding shared e-mobility systems: A multi-agent deep reinforcement learning approach}.
\newblock \bibinfo{journal}{IEEE Transactions on Intelligent Transportation Systems} \bibinfo{volume}{24}, \bibinfo{pages}{3868--3881}.
\bibitem[{Luo et~al.(2023b)Luo, Xu, Peng, Wang, Duan, Mahmood, Wen, Ding and Xu}]{luo2023aq2pnn}
\bibinfo{author}{Luo, Y.}, \bibinfo{author}{Xu, N.}, \bibinfo{author}{Peng, H.}, \bibinfo{author}{Wang, C.}, \bibinfo{author}{Duan, S.}, \bibinfo{author}{Mahmood, K.}, \bibinfo{author}{Wen, W.}, \bibinfo{author}{Ding, C.}, \bibinfo{author}{Xu, X.}, \bibinfo{year}{2023}b.
\newblock \bibinfo{title}{Aq2pnn: Enabling two-party privacy-preserving deep neural network inference with adaptive quantization}, in: \bibinfo{booktitle}{2023 56th IEEE/ACM International Symposium on Microarchitecture (MICRO)}, \bibinfo{organization}{IEEE}. pp. \bibinfo{pages}{628--640}.
\bibitem[{Luvizon et~al.(2022)Luvizon, Picard and Tabia}]{9luvizon2022consensus}
\bibinfo{author}{Luvizon, D.C.}, \bibinfo{author}{Picard, D.}, \bibinfo{author}{Tabia, H.}, \bibinfo{year}{2022}.
\newblock \bibinfo{title}{Consensus-based optimization for 3d human pose estimation in camera coordinates}.
\newblock \bibinfo{journal}{International Journal of Computer Vision} \bibinfo{volume}{130}, \bibinfo{pages}{869--882}.
\bibitem[{Ma et~al.(2021)Ma, Su, Wang, Ci and Wang}]{ma2021context}
\bibinfo{author}{Ma, X.}, \bibinfo{author}{Su, J.}, \bibinfo{author}{Wang, C.}, \bibinfo{author}{Ci, H.}, \bibinfo{author}{Wang, Y.}, \bibinfo{year}{2021}.
\newblock \bibinfo{title}{Context modeling in 3d human pose estimation: A unified perspective}, in: \bibinfo{booktitle}{Proceedings of the IEEE/CVF conference on computer vision and pattern recognition}, pp. \bibinfo{pages}{6238--6247}.
\bibitem[{Maaz et~al.(2022)Maaz, Shaker, Cholakkal, Khan, Zamir, Anwer and Shahbaz~Khan}]{29maaz2022edgenext}
\bibinfo{author}{Maaz, M.}, \bibinfo{author}{Shaker, A.}, \bibinfo{author}{Cholakkal, H.}, \bibinfo{author}{Khan, S.}, \bibinfo{author}{Zamir, S.W.}, \bibinfo{author}{Anwer, R.M.}, \bibinfo{author}{Shahbaz~Khan, F.}, \bibinfo{year}{2022}.
\newblock \bibinfo{title}{Edgenext: efficiently amalgamated cnn-transformer architecture for mobile vision applications}, in: \bibinfo{booktitle}{European conference on computer vision}, \bibinfo{organization}{Springer}. pp. \bibinfo{pages}{3--20}.
\bibitem[{Mart{\'\i}nez-Gonz{\'a}lez et~al.(2019)Mart{\'\i}nez-Gonz{\'a}lez, Villamizar, Can{\'e}vet and Odobez}]{martinez2019efficient}
\bibinfo{author}{Mart{\'\i}nez-Gonz{\'a}lez, A.}, \bibinfo{author}{Villamizar, M.}, \bibinfo{author}{Can{\'e}vet, O.}, \bibinfo{author}{Odobez, J.M.}, \bibinfo{year}{2019}.
\newblock \bibinfo{title}{Efficient convolutional neural networks for depth-based multi-person pose estimation}.
\newblock \bibinfo{journal}{IEEE Transactions on Circuits and Systems for Video Technology} \bibinfo{volume}{30}, \bibinfo{pages}{4207--4221}.
\bibitem[{Mehta et~al.(2017)Mehta, Rhodin, Casas, Fua, Sotnychenko, Xu and Theobalt}]{mehta2017monocular}
\bibinfo{author}{Mehta, D.}, \bibinfo{author}{Rhodin, H.}, \bibinfo{author}{Casas, D.}, \bibinfo{author}{Fua, P.}, \bibinfo{author}{Sotnychenko, O.}, \bibinfo{author}{Xu, W.}, \bibinfo{author}{Theobalt, C.}, \bibinfo{year}{2017}.
\newblock \bibinfo{title}{Monocular 3d human pose estimation in the wild using improved cnn supervision}, in: \bibinfo{booktitle}{2017 international conference on 3D vision (3DV)}, \bibinfo{organization}{IEEE}. pp. \bibinfo{pages}{506--516}.
\bibitem[{Mei(2023)}]{mei20233d}
\bibinfo{author}{Mei, Z.}, \bibinfo{year}{2023}.
\newblock \bibinfo{title}{3d image analysis of sports technical features and sports training methods based on artificial intelligence}.
\newblock \bibinfo{journal}{Journal of Testing and Evaluation} \bibinfo{volume}{51}, \bibinfo{pages}{189--200}.
\bibitem[{Nekoui et~al.(2020)Nekoui, Cruz and Cheng}]{nekoui2020falcons}
\bibinfo{author}{Nekoui, M.}, \bibinfo{author}{Cruz, F.O.T.}, \bibinfo{author}{Cheng, L.}, \bibinfo{year}{2020}.
\newblock \bibinfo{title}{Falcons: Fast learner-grader for contorted poses in sports}, in: \bibinfo{booktitle}{Proceedings of the IEEE/CVF conference on computer vision and pattern recognition workshops}, pp. \bibinfo{pages}{900--901}.
\bibitem[{Nogueira et~al.(2024)Nogueira, Oliveira and Teixeira}]{28nogueira2024markerless}
\bibinfo{author}{Nogueira, A.F.R.}, \bibinfo{author}{Oliveira, H.P.}, \bibinfo{author}{Teixeira, L.F.}, \bibinfo{year}{2024}.
\newblock \bibinfo{title}{Markerless multi-view 3d human pose estimation: a survey}.
\newblock \bibinfo{journal}{arXiv preprint arXiv:2407.03817} .
\bibitem[{Nowak and Lelowicz(2021a)}]{30nowak2021weight}
\bibinfo{author}{Nowak, M.K.}, \bibinfo{author}{Lelowicz, K.}, \bibinfo{year}{2021}a.
\newblock \bibinfo{title}{Weight perturbation as a method for improving performance of deep neural networks}, in: \bibinfo{booktitle}{2021 25th International Conference on Methods and Models in Automation and Robotics (MMAR)}, \bibinfo{organization}{IEEE}. pp. \bibinfo{pages}{127--132}.
\bibitem[{Nowak and Lelowicz(2021b)}]{31nowak2021weight}
\bibinfo{author}{Nowak, M.K.}, \bibinfo{author}{Lelowicz, K.}, \bibinfo{year}{2021}b.
\newblock \bibinfo{title}{Weight perturbation as a method for improving performance of deep neural networks}, in: \bibinfo{booktitle}{2021 25th International Conference on Methods and Models in Automation and Robotics (MMAR)}, \bibinfo{organization}{IEEE}. pp. \bibinfo{pages}{127--132}.
\bibitem[{Oikarinen et~al.(2021)Oikarinen, Hannah and Kazerounian}]{oikarinen2021graphmdn}
\bibinfo{author}{Oikarinen, T.}, \bibinfo{author}{Hannah, D.}, \bibinfo{author}{Kazerounian, S.}, \bibinfo{year}{2021}.
\newblock \bibinfo{title}{Graphmdn: Leveraging graph structure and deep learning to solve inverse problems}, in: \bibinfo{booktitle}{2021 International Joint Conference on Neural Networks (IJCNN)}, \bibinfo{organization}{IEEE}. pp. \bibinfo{pages}{1--9}.
\bibitem[{Patel et~al.(2020)Patel, Labana, Pandya, Modi, Ghayvat and Awais}]{patel2020histogram}
\bibinfo{author}{Patel, C.I.}, \bibinfo{author}{Labana, D.}, \bibinfo{author}{Pandya, S.}, \bibinfo{author}{Modi, K.}, \bibinfo{author}{Ghayvat, H.}, \bibinfo{author}{Awais, M.}, \bibinfo{year}{2020}.
\newblock \bibinfo{title}{Histogram of oriented gradient-based fusion of features for human action recognition in action video sequences}.
\newblock \bibinfo{journal}{Sensors} \bibinfo{volume}{20}, \bibinfo{pages}{7299}.
\bibitem[{Pavlakos et~al.(2018)Pavlakos, Zhou and Daniilidis}]{pavlakos2018ordinal}
\bibinfo{author}{Pavlakos, G.}, \bibinfo{author}{Zhou, X.}, \bibinfo{author}{Daniilidis, K.}, \bibinfo{year}{2018}.
\newblock \bibinfo{title}{Ordinal depth supervision for 3d human pose estimation}, in: \bibinfo{booktitle}{Proceedings of the IEEE conference on computer vision and pattern recognition}, pp. \bibinfo{pages}{7307--7316}.
\bibitem[{Pavllo et~al.(2019)Pavllo, Feichtenhofer, Grangier and Auli}]{pavllo20193d}
\bibinfo{author}{Pavllo, D.}, \bibinfo{author}{Feichtenhofer, C.}, \bibinfo{author}{Grangier, D.}, \bibinfo{author}{Auli, M.}, \bibinfo{year}{2019}.
\newblock \bibinfo{title}{3d human pose estimation in video with temporal convolutions and semi-supervised training}, in: \bibinfo{booktitle}{Proceedings of the IEEE/CVF conference on computer vision and pattern recognition}, pp. \bibinfo{pages}{7753--7762}.
\bibitem[{Peng et~al.(2023)Peng, Huang, Zhou, Luo, Wang, Wang, Zhao, Xie, Li, Geng et~al.}]{peng2023autorep}
\bibinfo{author}{Peng, H.}, \bibinfo{author}{Huang, S.}, \bibinfo{author}{Zhou, T.}, \bibinfo{author}{Luo, Y.}, \bibinfo{author}{Wang, C.}, \bibinfo{author}{Wang, Z.}, \bibinfo{author}{Zhao, J.}, \bibinfo{author}{Xie, X.}, \bibinfo{author}{Li, A.}, \bibinfo{author}{Geng, T.}, et~al., \bibinfo{year}{2023}.
\newblock \bibinfo{title}{Autorep: Automatic relu replacement for fast private network inference}, in: \bibinfo{booktitle}{2023 IEEE/CVF International Conference on Computer Vision (ICCV)}, \bibinfo{organization}{IEEE}. pp. \bibinfo{pages}{5155--5165}.
\bibitem[{Peng et~al.(2024a)Peng, Xie, Shivdikar, Hasan, Zhao, Huang, Khan, Kaeli and Ding}]{peng2024maxk}
\bibinfo{author}{Peng, H.}, \bibinfo{author}{Xie, X.}, \bibinfo{author}{Shivdikar, K.}, \bibinfo{author}{Hasan, M.A.}, \bibinfo{author}{Zhao, J.}, \bibinfo{author}{Huang, S.}, \bibinfo{author}{Khan, O.}, \bibinfo{author}{Kaeli, D.}, \bibinfo{author}{Ding, C.}, \bibinfo{year}{2024}a.
\newblock \bibinfo{title}{Maxk-gnn: Extremely fast gpu kernel design for accelerating graph neural networks training}, in: \bibinfo{booktitle}{Proceedings of the 29th ACM International Conference on Architectural Support for Programming Languages and Operating Systems, Volume 2}, \bibinfo{publisher}{Association for Computing Machinery}, \bibinfo{address}{New York, NY, USA}. p. \bibinfo{pages}{683–698}.
\bibitem[{Peng et~al.(2024b)Peng, Xu, Feng, Zhao, Tan, Zhou, Zhang, Gong and Zheng}]{peng2024automatic}
\bibinfo{author}{Peng, X.}, \bibinfo{author}{Xu, Q.}, \bibinfo{author}{Feng, Z.}, \bibinfo{author}{Zhao, H.}, \bibinfo{author}{Tan, L.}, \bibinfo{author}{Zhou, Y.}, \bibinfo{author}{Zhang, Z.}, \bibinfo{author}{Gong, C.}, \bibinfo{author}{Zheng, Y.}, \bibinfo{year}{2024}b.
\newblock \bibinfo{title}{Automatic news generation and fact-checking system based on language processing}.
\newblock \bibinfo{journal}{Journal of Industrial Engineering and Applied Science} \bibinfo{volume}{2}, \bibinfo{pages}{1--11}.
\bibitem[{Postolache et~al.(2020)Postolache, Hemanth, Alexandre, Gupta, Geman and Khanna}]{32postolache2020remote}
\bibinfo{author}{Postolache, O.}, \bibinfo{author}{Hemanth, D.J.}, \bibinfo{author}{Alexandre, R.}, \bibinfo{author}{Gupta, D.}, \bibinfo{author}{Geman, O.}, \bibinfo{author}{Khanna, A.}, \bibinfo{year}{2020}.
\newblock \bibinfo{title}{Remote monitoring of physical rehabilitation of stroke patients using iot and virtual reality}.
\newblock \bibinfo{journal}{IEEE Journal on Selected Areas in Communications} \bibinfo{volume}{39}, \bibinfo{pages}{562--573}.
\bibitem[{Qiao et~al.(2024)Qiao, Li, Lin, Wei, Jiang, Luo and Yang}]{qiao2024robust}
\bibinfo{author}{Qiao, Y.}, \bibinfo{author}{Li, K.}, \bibinfo{author}{Lin, J.}, \bibinfo{author}{Wei, R.}, \bibinfo{author}{Jiang, C.}, \bibinfo{author}{Luo, Y.}, \bibinfo{author}{Yang, H.}, \bibinfo{year}{2024}.
\newblock \bibinfo{title}{Robust domain generalization for multi-modal object recognition}, in: \bibinfo{booktitle}{2024 5th International Conference on Artificial Intelligence and Electromechanical Automation (AIEA)}, \bibinfo{organization}{IEEE}. pp. \bibinfo{pages}{392--397}.
\bibitem[{Ran et~al.(2024)Ran, Gao, Li, Li, Tian, Wang, Shi and Ning}]{ran2024brain}
\bibinfo{author}{Ran, H.}, \bibinfo{author}{Gao, X.}, \bibinfo{author}{Li, L.}, \bibinfo{author}{Li, W.}, \bibinfo{author}{Tian, S.}, \bibinfo{author}{Wang, G.}, \bibinfo{author}{Shi, H.}, \bibinfo{author}{Ning, X.}, \bibinfo{year}{2024}.
\newblock \bibinfo{title}{Brain-inspired fast-and slow-update prompt tuning for few-shot class-incremental learning}.
\newblock \bibinfo{journal}{IEEE Transactions on Neural Networks and Learning Systems} .
\bibitem[{Shan et~al.(2022)Shan, Liu, Zhang, Wang, Ma and Gao}]{shan2022p}
\bibinfo{author}{Shan, W.}, \bibinfo{author}{Liu, Z.}, \bibinfo{author}{Zhang, X.}, \bibinfo{author}{Wang, S.}, \bibinfo{author}{Ma, S.}, \bibinfo{author}{Gao, W.}, \bibinfo{year}{2022}.
\newblock \bibinfo{title}{P-stmo: Pre-trained spatial temporal many-to-one model for 3d human pose estimation}, in: \bibinfo{booktitle}{European Conference on Computer Vision}, \bibinfo{organization}{Springer}. pp. \bibinfo{pages}{461--478}.
\bibitem[{Sharma et~al.(2019)Sharma, Varigonda, Bindal, Sharma and Jain}]{sharma2019monocular}
\bibinfo{author}{Sharma, S.}, \bibinfo{author}{Varigonda, P.T.}, \bibinfo{author}{Bindal, P.}, \bibinfo{author}{Sharma, A.}, \bibinfo{author}{Jain, A.}, \bibinfo{year}{2019}.
\newblock \bibinfo{title}{Monocular 3d human pose estimation by generation and ordinal ranking}, in: \bibinfo{booktitle}{Proceedings of the IEEE/CVF international conference on computer vision}, pp. \bibinfo{pages}{2325--2334}.
\bibitem[{Shen et~al.(2024)Shen, Zhang, Zheng and Qi}]{Shen2024Harnessing}
\bibinfo{author}{Shen, X.}, \bibinfo{author}{Zhang, Q.}, \bibinfo{author}{Zheng, H.}, \bibinfo{author}{Qi, W.}, \bibinfo{year}{2024}.
\newblock \bibinfo{title}{{Harnessing XGBoost for robust biomarker selection of obsessive-compulsive disorder (OCD) from adolescent brain cognitive development (ABCD) data}}, in: \bibinfo{editor}{Piccaluga, P.P.}, \bibinfo{editor}{El-Hashash, A.}, \bibinfo{editor}{Guo, X.} (Eds.), \bibinfo{booktitle}{Fourth International Conference on Biomedicine and Bioinformatics Engineering (ICBBE 2024)}, \bibinfo{organization}{International Society for Optics and Photonics}. \bibinfo{publisher}{SPIE}. p. \bibinfo{pages}{132520U}.
\newblock \URLprefix \url{https://doi.org/10.1117/12.3044221}, \DOIprefix\doi{10.1117/12.3044221}.
\bibitem[{Sigal et~al.(2010)Sigal, Balan and Black}]{sigal2010humaneva}
\bibinfo{author}{Sigal, L.}, \bibinfo{author}{Balan, A.O.}, \bibinfo{author}{Black, M.J.}, \bibinfo{year}{2010}.
\newblock \bibinfo{title}{Humaneva: Synchronized video and motion capture dataset and baseline algorithm for evaluation of articulated human motion}.
\newblock \bibinfo{journal}{International journal of computer vision} \bibinfo{volume}{87}, \bibinfo{pages}{4--27}.
\bibitem[{Song et~al.(2024)Song, Li, Chen and Demachi}]{song2024quater}
\bibinfo{author}{Song, X.}, \bibinfo{author}{Li, Z.}, \bibinfo{author}{Chen, S.}, \bibinfo{author}{Demachi, K.}, \bibinfo{year}{2024}.
\newblock \bibinfo{title}{Quater-gcn: Enhancing 3d human pose estimation with orientation and semi-supervised training}.
\newblock \bibinfo{journal}{arXiv preprint arXiv:2404.19279} .
\bibitem[{Sui et~al.(2024)Sui, Jiang, Lyu, Wang, Zhou, Chen and Alhosain}]{sui2024application}
\bibinfo{author}{Sui, M.}, \bibinfo{author}{Jiang, L.}, \bibinfo{author}{Lyu, T.}, \bibinfo{author}{Wang, H.}, \bibinfo{author}{Zhou, L.}, \bibinfo{author}{Chen, P.}, \bibinfo{author}{Alhosain, A.}, \bibinfo{year}{2024}.
\newblock \bibinfo{title}{Application of deep learning models based on efficientdet and openpose in user-oriented motion rehabilitation robot control}.
\newblock \bibinfo{journal}{Journal of Intelligence Technology and Innovation} \bibinfo{volume}{2}, \bibinfo{pages}{47--77}.
\bibitem[{Sun et~al.(2020a)Sun, Wang, Zhao and Zhang}]{sun2020multi}
\bibinfo{author}{Sun, J.}, \bibinfo{author}{Wang, M.}, \bibinfo{author}{Zhao, X.}, \bibinfo{author}{Zhang, D.}, \bibinfo{year}{2020}a.
\newblock \bibinfo{title}{Multi-view pose generator based on deep learning for monocular 3d human pose estimation}.
\newblock \bibinfo{journal}{Symmetry} \bibinfo{volume}{12}, \bibinfo{pages}{1116}.
\bibitem[{Sun et~al.(2020b)Sun, Wang, Zhao and Zhang}]{10sun2020multi}
\bibinfo{author}{Sun, J.}, \bibinfo{author}{Wang, M.}, \bibinfo{author}{Zhao, X.}, \bibinfo{author}{Zhang, D.}, \bibinfo{year}{2020}b.
\newblock \bibinfo{title}{Multi-view pose generator based on deep learning for monocular 3d human pose estimation}.
\newblock \bibinfo{journal}{Symmetry} \bibinfo{volume}{12}, \bibinfo{pages}{1116}.
\bibitem[{Wan et~al.(2024)Wan, Zhang, Jiang, Wang and Zhou}]{wan2024image}
\bibinfo{author}{Wan, Q.}, \bibinfo{author}{Zhang, Z.}, \bibinfo{author}{Jiang, L.}, \bibinfo{author}{Wang, Z.}, \bibinfo{author}{Zhou, Y.}, \bibinfo{year}{2024}.
\newblock \bibinfo{title}{Image anomaly detection and prediction scheme based on ssa optimized resnet50-bigru model}.
\newblock \bibinfo{journal}{arXiv preprint arXiv:2406.13987} .
\bibitem[{Wang et~al.(2024a)Wang, Sui, Sun, Zhang and Zhou}]{Wang2024Theoretical}
\bibinfo{author}{Wang, C.}, \bibinfo{author}{Sui, M.}, \bibinfo{author}{Sun, D.}, \bibinfo{author}{Zhang, Z.}, \bibinfo{author}{Zhou, Y.}, \bibinfo{year}{2024}a.
\newblock \bibinfo{title}{Theoretical analysis of meta reinforcement learning: Generalization bounds and convergence guarantees}, \bibinfo{publisher}{Association for Computing Machinery}, \bibinfo{address}{New York, NY, USA}. p. \bibinfo{pages}{153–159}.
\newblock \URLprefix \url{https://doi.org/10.1145/3677779.3677804}, \DOIprefix\doi{10.1145/3677779.3677804}.
\bibitem[{Wang et~al.(2019)Wang, Huang, Wang and Tao}]{wang2019not}
\bibinfo{author}{Wang, J.}, \bibinfo{author}{Huang, S.}, \bibinfo{author}{Wang, X.}, \bibinfo{author}{Tao, D.}, \bibinfo{year}{2019}.
\newblock \bibinfo{title}{Not all parts are created equal: 3d human pose estimation by modeling bi-directional dependencies of body parts}, in: \bibinfo{booktitle}{International Conference on Computer Vision 2019}.
\bibitem[{Wang et~al.(2021)Wang, Tan, Zhen, Xu, Zheng, He and Shao}]{3wang2021deep}
\bibinfo{author}{Wang, J.}, \bibinfo{author}{Tan, S.}, \bibinfo{author}{Zhen, X.}, \bibinfo{author}{Xu, S.}, \bibinfo{author}{Zheng, F.}, \bibinfo{author}{He, Z.}, \bibinfo{author}{Shao, L.}, \bibinfo{year}{2021}.
\newblock \bibinfo{title}{Deep 3d human pose estimation: A review}.
\newblock \bibinfo{journal}{Computer Vision and Image Understanding} \bibinfo{volume}{210}, \bibinfo{pages}{103225}.
\bibitem[{Wang et~al.(2024b)Wang, Wang and Liu}]{wang2024recording}
\bibinfo{author}{Wang, J.}, \bibinfo{author}{Wang, Z.}, \bibinfo{author}{Liu, G.}, \bibinfo{year}{2024}b.
\newblock \bibinfo{title}{Recording brain activity while listening to music using wearable eeg devices combined with bidirectional long short-term memory networks}.
\newblock \bibinfo{journal}{Alexandria Engineering Journal} \bibinfo{volume}{109}, \bibinfo{pages}{1--10}.
\bibitem[{Wang et~al.(2020)Wang, Yan, Xiong and Lin}]{wang2020motion}
\bibinfo{author}{Wang, J.}, \bibinfo{author}{Yan, S.}, \bibinfo{author}{Xiong, Y.}, \bibinfo{author}{Lin, D.}, \bibinfo{year}{2020}.
\newblock \bibinfo{title}{Motion guided 3d pose estimation from videos}, in: \bibinfo{booktitle}{European conference on computer vision}, \bibinfo{organization}{Springer}. pp. \bibinfo{pages}{764--780}.
\bibitem[{Wang et~al.(2024c)Wang, Hu and Zhou}]{wang2024cross}
\bibinfo{author}{Wang, L.}, \bibinfo{author}{Hu, Y.}, \bibinfo{author}{Zhou, Y.}, \bibinfo{year}{2024}c.
\newblock \bibinfo{title}{Cross-border commodity pricing strategy optimization via mixed neural network for time series analysis}.
\newblock \bibinfo{journal}{arXiv preprint arXiv:2408.12115} .
\bibitem[{Wang et~al.(2024d)Wang, Ji, Wang, Feng and Du}]{wang2024intelligent}
\bibinfo{author}{Wang, L.}, \bibinfo{author}{Ji, W.}, \bibinfo{author}{Wang, G.}, \bibinfo{author}{Feng, Y.}, \bibinfo{author}{Du, M.}, \bibinfo{year}{2024}d.
\newblock \bibinfo{title}{Intelligent design and optimization of exercise equipment based on fusion algorithm of yolov5-resnet 50}.
\newblock \bibinfo{journal}{Alexandria Engineering Journal} \bibinfo{volume}{104}, \bibinfo{pages}{710--722}.
\bibitem[{Wang et~al.(2024e)Wang, Jiang, Wang and Zhou}]{wang2024deep}
\bibinfo{author}{Wang, S.}, \bibinfo{author}{Jiang, R.}, \bibinfo{author}{Wang, Z.}, \bibinfo{author}{Zhou, Y.}, \bibinfo{year}{2024}e.
\newblock \bibinfo{title}{Deep learning-based anomaly detection and log analysis for computer networks}.
\newblock \bibinfo{journal}{Journal of Information and Computing} \bibinfo{volume}{2}, \bibinfo{pages}{34--63}.
\bibitem[{Wang et~al.(2018)Wang, Zhao, Li, Fraire, Sun and Fang}]{wang2018performance}
\bibinfo{author}{Wang, Y.}, \bibinfo{author}{Zhao, K.}, \bibinfo{author}{Li, W.}, \bibinfo{author}{Fraire, J.}, \bibinfo{author}{Sun, Z.}, \bibinfo{author}{Fang, Y.}, \bibinfo{year}{2018}.
\newblock \bibinfo{title}{Performance evaluation of quic with bbr in satellite internet}, in: \bibinfo{booktitle}{2018 6th IEEE International Conference on Wireless for Space and Extreme Environments (WiSEE)}, \bibinfo{organization}{IEEE}. pp. \bibinfo{pages}{195--199}.
\bibitem[{Wehrbein et~al.(2021)Wehrbein, Rudolph, Rosenhahn and Wandt}]{wehrbein2021probabilistic}
\bibinfo{author}{Wehrbein, T.}, \bibinfo{author}{Rudolph, M.}, \bibinfo{author}{Rosenhahn, B.}, \bibinfo{author}{Wandt, B.}, \bibinfo{year}{2021}.
\newblock \bibinfo{title}{Probabilistic monocular 3d human pose estimation with normalizing flows}, in: \bibinfo{booktitle}{Proceedings of the IEEE/CVF international conference on computer vision}, pp. \bibinfo{pages}{11199--11208}.
\bibitem[{Weng(2024)}]{weng2024big}
\bibinfo{author}{Weng, Y.}, \bibinfo{year}{2024}.
\newblock \bibinfo{title}{Big data and machine learning in defence}.
\newblock \bibinfo{journal}{International Journal of Computer Science and Information Technology} \bibinfo{volume}{16}, \bibinfo{pages}{25--35}.
\bibitem[{Weng and Wu(2024)}]{weng2024leveraging}
\bibinfo{author}{Weng, Y.}, \bibinfo{author}{Wu, J.}, \bibinfo{year}{2024}.
\newblock \bibinfo{title}{Leveraging artificial intelligence to enhance data security and combat cyber attacks}.
\newblock \bibinfo{journal}{Journal of Artificial Intelligence General science (JAIGS) ISSN: 3006-4023} \bibinfo{volume}{5}, \bibinfo{pages}{392--399}.
\newblock \DOIprefix\doi{10.60087/jaigs.v5i1.211}.
\bibitem[{Weng et~al.(2024)Weng, Wu et~al.}]{weng2024fortifying}
\bibinfo{author}{Weng, Y.}, \bibinfo{author}{Wu, J.}, et~al., \bibinfo{year}{2024}.
\newblock \bibinfo{title}{Fortifying the global data fortress: a multidimensional examination of cyber security indexes and data protection measures across 193 nations}.
\newblock \bibinfo{journal}{International Journal of Frontiers in Engineering Technology} \bibinfo{volume}{6}, \bibinfo{pages}{13--28}.
\bibitem[{Xi et~al.(2024)Xi, Zhang, Jia and Jiang}]{xi2024enhancing}
\bibinfo{author}{Xi, X.}, \bibinfo{author}{Zhang, C.}, \bibinfo{author}{Jia, W.}, \bibinfo{author}{Jiang, R.}, \bibinfo{year}{2024}.
\newblock \bibinfo{title}{Enhancing human pose estimation in sports training: Integrating spatiotemporal transformer for improved accuracy and real-time performance}.
\newblock \bibinfo{journal}{Alexandria Engineering Journal} \bibinfo{volume}{109}, \bibinfo{pages}{144--156}.
\bibitem[{Xia and Xiao(2020)}]{27xia20203d}
\bibinfo{author}{Xia, H.}, \bibinfo{author}{Xiao, M.}, \bibinfo{year}{2020}.
\newblock \bibinfo{title}{3d human pose estimation with generative adversarial networks}.
\newblock \bibinfo{journal}{IEEE Access} \bibinfo{volume}{8}, \bibinfo{pages}{206198--206206}.
\bibitem[{Xu et~al.(2020)Xu, Yu, Ni, Yang, Yang and Zhang}]{xu2020deep}
\bibinfo{author}{Xu, J.}, \bibinfo{author}{Yu, Z.}, \bibinfo{author}{Ni, B.}, \bibinfo{author}{Yang, J.}, \bibinfo{author}{Yang, X.}, \bibinfo{author}{Zhang, W.}, \bibinfo{year}{2020}.
\newblock \bibinfo{title}{Deep kinematics analysis for monocular 3d human pose estimation}, in: \bibinfo{booktitle}{Proceedings of the IEEE/CVF Conference on computer vision and Pattern recognition}, pp. \bibinfo{pages}{899--908}.
\bibitem[{Xu and Takano(2021)}]{xu2021graph}
\bibinfo{author}{Xu, T.}, \bibinfo{author}{Takano, W.}, \bibinfo{year}{2021}.
\newblock \bibinfo{title}{Graph stacked hourglass networks for 3d human pose estimation}, in: \bibinfo{booktitle}{Proceedings of the IEEE/CVF conference on computer vision and pattern recognition}, pp. \bibinfo{pages}{16105--16114}.
\bibitem[{Xu et~al.(2022a)Xu, Zhang, Zhang and Tao}]{xu2022vitpose}
\bibinfo{author}{Xu, Y.}, \bibinfo{author}{Zhang, J.}, \bibinfo{author}{Zhang, Q.}, \bibinfo{author}{Tao, D.}, \bibinfo{year}{2022}a.
\newblock \bibinfo{title}{Vitpose: Simple vision transformer baselines for human pose estimation}.
\newblock \bibinfo{journal}{Advances in Neural Information Processing Systems} \bibinfo{volume}{35}, \bibinfo{pages}{38571--38584}.
\bibitem[{Xu et~al.(2022b)Xu, Deng, Dong and Shimada}]{xu2022dpmpc}
\bibinfo{author}{Xu, Z.}, \bibinfo{author}{Deng, D.}, \bibinfo{author}{Dong, Y.}, \bibinfo{author}{Shimada, K.}, \bibinfo{year}{2022}b.
\newblock \bibinfo{title}{Dpmpc-planner: A real-time uav trajectory planning framework for complex static environments with dynamic obstacles}, in: \bibinfo{booktitle}{2022 International Conference on Robotics and Automation (ICRA)}, \bibinfo{organization}{IEEE}. pp. \bibinfo{pages}{250--256}.
\bibitem[{Yan et~al.(2024)Yan, Wu, Kumar and Zhou}]{yan2024application}
\bibinfo{author}{Yan, T.}, \bibinfo{author}{Wu, J.}, \bibinfo{author}{Kumar, M.}, \bibinfo{author}{Zhou, Y.}, \bibinfo{year}{2024}.
\newblock \bibinfo{title}{Application of deep learning for automatic identification of hazardous materials and urban safety supervision}.
\newblock \bibinfo{journal}{Journal of Organizational and End User Computing (JOEUC)} \bibinfo{volume}{36}, \bibinfo{pages}{1--20}.
\bibitem[{Yang et~al.(2018)Yang, Ouyang, Wang, Ren, Li and Wang}]{yang20183d}
\bibinfo{author}{Yang, W.}, \bibinfo{author}{Ouyang, W.}, \bibinfo{author}{Wang, X.}, \bibinfo{author}{Ren, J.}, \bibinfo{author}{Li, H.}, \bibinfo{author}{Wang, X.}, \bibinfo{year}{2018}.
\newblock \bibinfo{title}{3d human pose estimation in the wild by adversarial learning}, in: \bibinfo{booktitle}{Proceedings of the IEEE conference on computer vision and pattern recognition}, pp. \bibinfo{pages}{5255--5264}.
\bibitem[{Yang et~al.(2016)Yang, Yang, Dyer, He, Smola and Hovy}]{yang2016hierarchical}
\bibinfo{author}{Yang, Z.}, \bibinfo{author}{Yang, D.}, \bibinfo{author}{Dyer, C.}, \bibinfo{author}{He, X.}, \bibinfo{author}{Smola, A.}, \bibinfo{author}{Hovy, E.}, \bibinfo{year}{2016}.
\newblock \bibinfo{title}{Hierarchical attention networks for document classification}, in: \bibinfo{booktitle}{Proceedings of the 2016 conference of the North American chapter of the association for computational linguistics: human language technologies}, pp. \bibinfo{pages}{1480--1489}.
\bibitem[{Yu et~al.(2023)Yu, Zhang, Liu, Zhong, Liu and Chen}]{yu2023gla}
\bibinfo{author}{Yu, B.X.}, \bibinfo{author}{Zhang, Z.}, \bibinfo{author}{Liu, Y.}, \bibinfo{author}{Zhong, S.h.}, \bibinfo{author}{Liu, Y.}, \bibinfo{author}{Chen, C.W.}, \bibinfo{year}{2023}.
\newblock \bibinfo{title}{Gla-gcn: Global-local adaptive graph convolutional network for 3d human}.
\newblock \bibinfo{journal}{arXiv preprint arXiv:2307.05853} .
\bibitem[{Zeng et~al.(2020)Zeng, Sun, Huang, Liu, Xu and Lin}]{zeng2020srnet}
\bibinfo{author}{Zeng, A.}, \bibinfo{author}{Sun, X.}, \bibinfo{author}{Huang, F.}, \bibinfo{author}{Liu, M.}, \bibinfo{author}{Xu, Q.}, \bibinfo{author}{Lin, S.}, \bibinfo{year}{2020}.
\newblock \bibinfo{title}{Srnet: Improving generalization in 3d human pose estimation with a split-and-recombine approach}, in: \bibinfo{booktitle}{Computer Vision--ECCV 2020: 16th European Conference, Glasgow, UK, August 23--28, 2020, Proceedings, Part XIV 16}, \bibinfo{organization}{Springer}. pp. \bibinfo{pages}{507--523}.
\bibitem[{Zhang et~al.(2022a)Zhang, Lu, Zhan and Zhang}]{6zhang2022semi}
\bibinfo{author}{Zhang, H.}, \bibinfo{author}{Lu, G.}, \bibinfo{author}{Zhan, M.}, \bibinfo{author}{Zhang, B.}, \bibinfo{year}{2022}a.
\newblock \bibinfo{title}{Semi-supervised classification of graph convolutional networks with laplacian rank constraints}.
\newblock \bibinfo{journal}{Neural Processing Letters} , \bibinfo{pages}{1--12}.
\bibitem[{Zhang and Tao(2020)}]{zhang2020empowering}
\bibinfo{author}{Zhang, J.}, \bibinfo{author}{Tao, D.}, \bibinfo{year}{2020}.
\newblock \bibinfo{title}{Empowering things with intelligence: a survey of the progress, challenges, and opportunities in artificial intelligence of things}.
\newblock \bibinfo{journal}{IEEE Internet of Things Journal} \bibinfo{volume}{8}, \bibinfo{pages}{7789--7817}.
\bibitem[{Zhang et~al.(2022b)Zhang, Tu, Yang, Chen and Yuan}]{zhang2022mixste}
\bibinfo{author}{Zhang, J.}, \bibinfo{author}{Tu, Z.}, \bibinfo{author}{Yang, J.}, \bibinfo{author}{Chen, Y.}, \bibinfo{author}{Yuan, J.}, \bibinfo{year}{2022}b.
\newblock \bibinfo{title}{Mixste: Seq2seq mixed spatio-temporal encoder for 3d human pose estimation in video}, in: \bibinfo{booktitle}{Proceedings of the IEEE/CVF conference on computer vision and pattern recognition}, pp. \bibinfo{pages}{13232--13242}.
\bibitem[{Zhang et~al.(2024)Zhang, Qi, Zheng and Shen}]{zhang2024cu}
\bibinfo{author}{Zhang, Q.}, \bibinfo{author}{Qi, W.}, \bibinfo{author}{Zheng, H.}, \bibinfo{author}{Shen, X.}, \bibinfo{year}{2024}.
\newblock \bibinfo{title}{Cu-net: a u-net architecture for efficient brain-tumor segmentation on brats 2019 dataset}.
\newblock \bibinfo{journal}{arXiv preprint arXiv:2406.13113} .
\bibitem[{Zhang et~al.(2023)Zhang, Ji, Wang, Mei, Kortylewski and Yuille}]{7zhang20233d}
\bibinfo{author}{Zhang, Y.}, \bibinfo{author}{Ji, P.}, \bibinfo{author}{Wang, A.}, \bibinfo{author}{Mei, J.}, \bibinfo{author}{Kortylewski, A.}, \bibinfo{author}{Yuille, A.}, \bibinfo{year}{2023}.
\newblock \bibinfo{title}{3d-aware neural body fitting for occlusion robust 3d human pose estimation}, in: \bibinfo{booktitle}{Proceedings of the IEEE/CVF International Conference on Computer Vision}, pp. \bibinfo{pages}{9399--9410}.
\bibitem[{Zhang(2024)}]{zhang2024deep}
\bibinfo{author}{Zhang, Z.}, \bibinfo{year}{2024}.
\newblock \bibinfo{title}{Deep analysis of time series data for smart grid startup strategies: A transformer-lstm-pso model approach}.
\newblock \bibinfo{journal}{Journal of Management Science and Operations} \bibinfo{volume}{2}, \bibinfo{pages}{16--43}.
\bibitem[{Zhao et~al.(2022)Zhao, Wang and Tian}]{zhao2022graformer}
\bibinfo{author}{Zhao, W.}, \bibinfo{author}{Wang, W.}, \bibinfo{author}{Tian, Y.}, \bibinfo{year}{2022}.
\newblock \bibinfo{title}{Graformer: Graph-oriented transformer for 3d pose estimation}, in: \bibinfo{booktitle}{Proceedings of the IEEE/CVF Conference on Computer Vision and Pattern Recognition}, pp. \bibinfo{pages}{20438--20447}.
\bibitem[{Zheng et~al.(2023)Zheng, Wu, Chen, Yang, Zhu, Shen, Kehtarnavaz and Shah}]{4zheng2023deep}
\bibinfo{author}{Zheng, C.}, \bibinfo{author}{Wu, W.}, \bibinfo{author}{Chen, C.}, \bibinfo{author}{Yang, T.}, \bibinfo{author}{Zhu, S.}, \bibinfo{author}{Shen, J.}, \bibinfo{author}{Kehtarnavaz, N.}, \bibinfo{author}{Shah, M.}, \bibinfo{year}{2023}.
\newblock \bibinfo{title}{Deep learning-based human pose estimation: A survey}.
\newblock \bibinfo{journal}{ACM Computing Surveys} \bibinfo{volume}{56}, \bibinfo{pages}{1--37}.
\bibitem[{Zheng et~al.(2021)Zheng, Zhu, Mendieta, Yang, Chen and Ding}]{zheng20213d}
\bibinfo{author}{Zheng, C.}, \bibinfo{author}{Zhu, S.}, \bibinfo{author}{Mendieta, M.}, \bibinfo{author}{Yang, T.}, \bibinfo{author}{Chen, C.}, \bibinfo{author}{Ding, Z.}, \bibinfo{year}{2021}.
\newblock \bibinfo{title}{3d human pose estimation with spatial and temporal transformers}, in: \bibinfo{booktitle}{Proceedings of the IEEE/CVF international conference on computer vision}, pp. \bibinfo{pages}{11656--11665}.
\bibitem[{Zheng et~al.(2024)Zheng, Zhang, Gong, Liu and Chen}]{zheng2024identification}
\bibinfo{author}{Zheng, H.}, \bibinfo{author}{Zhang, Q.}, \bibinfo{author}{Gong, Y.}, \bibinfo{author}{Liu, Z.}, \bibinfo{author}{Chen, S.}, \bibinfo{year}{2024}.
\newblock \bibinfo{title}{Identification of prognostic biomarkers for stage iii non-small cell lung carcinoma in female nonsmokers using machine learning}.
\newblock \bibinfo{journal}{arXiv preprint arXiv:2408.16068} .
\bibitem[{Zhou et~al.(2021)Zhou, Han, Jiang, Jia and Lu}]{zhou2021hemlets}
\bibinfo{author}{Zhou, K.}, \bibinfo{author}{Han, X.}, \bibinfo{author}{Jiang, N.}, \bibinfo{author}{Jia, K.}, \bibinfo{author}{Lu, J.}, \bibinfo{year}{2021}.
\newblock \bibinfo{title}{Hemlets posh: Learning part-centric heatmap triplets for 3d human pose and shape estimation}.
\newblock \bibinfo{journal}{IEEE Transactions on Pattern Analysis and Machine Intelligence} \bibinfo{volume}{44}, \bibinfo{pages}{3000--3014}.
\bibitem[{Zhou et~al.(2024a)Zhou, Zhao, Luo, Xie, Wen, Ding and Xu}]{zhou2024adapi}
\bibinfo{author}{Zhou, T.}, \bibinfo{author}{Zhao, J.}, \bibinfo{author}{Luo, Y.}, \bibinfo{author}{Xie, X.}, \bibinfo{author}{Wen, W.}, \bibinfo{author}{Ding, C.}, \bibinfo{author}{Xu, X.}, \bibinfo{year}{2024}a.
\newblock \bibinfo{title}{Adapi: Facilitating dnn model adaptivity for efficient private inference in edge computing}.
\newblock \bibinfo{journal}{arXiv preprint arXiv:2407.05633} .
\bibitem[{Zhou et~al.(2020)Zhou, Koltun and Kr{\"a}henb{\"u}hl}]{24zhou2020tracking}
\bibinfo{author}{Zhou, X.}, \bibinfo{author}{Koltun, V.}, \bibinfo{author}{Kr{\"a}henb{\"u}hl, P.}, \bibinfo{year}{2020}.
\newblock \bibinfo{title}{Tracking objects as points}, in: \bibinfo{booktitle}{European conference on computer vision}, \bibinfo{organization}{Springer}. pp. \bibinfo{pages}{474--490}.
\bibitem[{Zhou et~al.(2024b)Zhou, Wang, Zheng, Zhou, Dai, Luo, Zhang and Sui}]{zhou2024optimization}
\bibinfo{author}{Zhou, Y.}, \bibinfo{author}{Wang, Z.}, \bibinfo{author}{Zheng, S.}, \bibinfo{author}{Zhou, L.}, \bibinfo{author}{Dai, L.}, \bibinfo{author}{Luo, H.}, \bibinfo{author}{Zhang, Z.}, \bibinfo{author}{Sui, M.}, \bibinfo{year}{2024}b.
\newblock \bibinfo{title}{Optimization of automated garbage recognition model based on resnet-50 and weakly supervised cnn for sustainable urban development}.
\newblock \bibinfo{journal}{Alexandria Engineering Journal} \bibinfo{volume}{108}, \bibinfo{pages}{415--427}.

\end{thebibliography}


\end{document}